%% file: main.tex
\documentclass[lettersize,journal]{IEEEtran}
\usepackage{amsmath,amsfonts}
\usepackage{algorithmic}
\usepackage{algorithm}
\usepackage{array}
\usepackage[caption=false,font=normalsize,labelfont=sf,textfont=sf]{subfig}
\usepackage{textcomp}
\usepackage{stfloats}
\usepackage{url}
\usepackage{verbatim}
\usepackage{graphicx}
\usepackage{cite}
\usepackage[acronym,nonumberlist]{glossaries}
\input{acronyms.tex}
\usepackage{mathtools}
\usepackage{amsthm}
\theoremstyle{definition}
\newtheorem{definition}{Definition}[section]
\usepackage{stfloats}

\theoremstyle{lemma}

\begin{document}

\title{Motion Planning for Autonomous Vehicles using Optimization over Graphs of Convex Sets }

\author{\IEEEauthorblockN{Matheus Wagner, Antônio Augusto Fröhlich} \\
  \IEEEauthorblockA{\textit{Software/Hardware Integration Lab}, \textit{Federal University of Santa Catarina}, Florianópolis, Brazil \\
    \{wagner,guto\}@lisha.ufsc.br}
}

\markboth{Journal of \LaTeX\ Class Files,~Vol.~14, No.~8, August~2021}%
{Shell \MakeLowercase{\textit{et al.}}: A Sample Article Using IEEEtran.cls for IEEE Journals}


\maketitle
\begin{abstract}
Motion planning for autonomous vehicles requires generating collision-free and dynamically feasible trajectories in complex environments under real-time constraints. While nonlinear optimal control formulations provide high-fidelity solutions, they are computationally demanding and sensitive to initialization, whereas geometric planning methods scale well but often decouple path selection from trajectory optimization. This paper studies the extent to which optimization over Graphs of Convex Sets (GCS) can approximate solutions of nonlinear optimal control problems in the context of autonomous driving. The free space is represented as a finite union of convex regions organized as a directed graph, allowing nonconvex geometry to be handled through discrete connectivity decisions while maintaining convex trajectory constraints within each region. Vehicle motion is parameterized using Bezier curves for the spatial path and a polynomial time-scaling function for temporal evolution. Under small-slip and linear tire assumptions, a simplified dynamic bicycle model enables approximate enforcement of dynamic feasibility through convex constraints on trajectory derivatives. The approach is evaluated in CommonRoad scenarios involving static obstacle avoidance and lane-changing maneuvers, and is compared against a nonlinear discrete-time optimal control formulation. The results indicate that the GCS-based method generates collision-free and dynamically consistent trajectories that closely match those obtained from the nonlinear program, while exhibiting improved computational efficiency and reduced sensitivity to initialization. These findings suggest that GCS provides a structured approximation of nonlinear motion planning problems, capturing dominant geometric and dynamic effects while preserving convexity in the continuous relaxation.
\end{abstract}


\section{Introduction}

Autonomous driving requires motion planning algorithms capable of generating collision-free and dynamically feasible trajectories in geometrically complex, uncertain, and time-varying environments. Although substantial advances have been achieved in perception, control, and learning-based components of autonomous systems \cite{yu2021model, badue2021self}, ensuring safe and reliable decision-making under strict real-time constraints remains challenging. Motion planning plays a central role in this architecture, as it must reconcile vehicle dynamics, obstacle avoidance, traffic rules, and computational tractability within a unified framework \cite{Hu2025ARO, paden2016survey}.

The motion planning problem consists of determining a trajectory that connects a start and a goal configuration while satisfying kinematic and dynamic constraints and avoiding collisions \cite{botros2023Spatio, paden2016survey}. Obstacle avoidance induces a nonconvex admissible configuration set, while vehicle dynamics introduce nonlinear constraints, increasing problem difficulty.

Existing approaches can be broadly categorized as search-based, sampling-based, optimization-based, and learning-based methods \cite{botros2023Spatio, teng2023motion, paden2016survey}. Search-based planners discretize the configuration space and rely on motion primitives, often limiting optimality and smoothness \cite{teng2023motion}. Sampling-based planners such as \gls{RRT} and its variants \cite{lavalle1998rapidly, yu2024rdt, orthey2023sampling} provide probabilistic completeness and scale well to high-dimensional spaces, but typically require post-processing to ensure dynamic feasibility. Learning-based methods leverage data-driven models to generate trajectories \cite{fan2024risk, wang2020learning, yang2024diffusion}, but generally lack formal safety guarantees and remain sensitive to distributional shift \cite{zhu2025Diff}.

Optimization-based methods encode vehicle models, safety constraints, and performance objectives within a unified mathematical formulation \cite{zhu2025Diff, betts2010practical}. These include optimal control formulations, which lead to nonconvex \gls{NLP}s solved via direct transcription and numerical optimization \cite{micheli2023nmpc, wagner2023rti}, and parametric trajectory representations based on curves such as Bézier curves or clothoids \cite{chen2014quartic, alia2015local, zayou2025Graph}, which enable dimensionality reduction and direct enforcement of smoothness. However, obstacle avoidance remains nonconvex, and many approaches rely on local methods such as iterative linearization or sequential convex programming \cite{zhu2025Diff, wang2021path}. While computationally efficient, these methods do not capture the combinatorial structure associated with alternative homotopy classes and are often sensitive to initialization.

An alternative perspective is provided by optimization over \gls{GCS} \cite{marcucci2023motion, marcucci2025unified}, where the free space is decomposed into convex regions organized as a graph. Motion planning is then formulated as a mixed-integer convex optimization problem in which binary variables select a sequence of regions while continuous variables describe trajectories within them. This formulation represents geometric nonconvexity through graph connectivity while preserving convexity of trajectory constraints within each region. In contrast to methods based on precomputed safe corridors \cite{zayou2025Graph}, \gls{GCS} integrates region selection and trajectory optimization within a single optimization problem.

Nevertheless, \gls{GCS} was originally developed for general motion planning and does not directly account for structural aspects specific to autonomous driving. Vehicle models introduce kinematic and dynamic constraints, and dynamic obstacles induce time-dependent modifications of feasible regions. Incorporating these elements requires careful modeling to preserve convexity while maintaining computational tractability.

A central question addressed in this work is the extent to which a \gls{GCS}-based formulation can reproduce solutions of the corresponding nonlinear optimal control problem. In this context, approximation is understood in terms of trajectory geometry, dynamic profiles, and feasibility with respect to vehicle constraints, rather than strict optimality guarantees. This perspective positions the proposed approach as a structured relaxation rather than a direct substitute for nonlinear optimal control.

In this work, the \gls{GCS} framework is specialized to autonomous driving by incorporating approximate vehicle dynamics and introducing a structured mechanism to account for dynamic obstacles within the graph-based formulation. The goal is to assess its ability to approximate trajectories obtained from nonlinear optimal control formulations. While the results are encouraging, the approach should be regarded as an initial investigation whose theoretical and practical properties require further analysis.

The main elements explored in this study are summarized as follows:
\begin{itemize}
\item A formulation of motion planning using optimization over \gls{GCS} that incorporates approximate vehicle dynamics through differential flatness, enabling convex constraints on dynamically relevant quantities.
\item An analysis of the extent to which \gls{GCS}-based trajectories approximate those obtained from a nonlinear optimal control formulation, across multiple driving scenarios.
\item A structured, though heuristic, mechanism for incorporating time-dependent obstacle avoidance within the \gls{GCS} framework, highlighting both its applicability and limitations.
\end{itemize}

\section{Preliminaries}

This section introduces the main modeling and mathematical tools used in the proposed formulation. The motion planning problem considered in this work combines geometric reasoning, vehicle dynamics, and optimization over structured representations of the free space. To this end, we first present the vehicle model and the assumptions used to relate the system dynamics to the trajectory in the geometric space. We then introduce the polynomial trajectory representation and the \gls{GCS} framework, which together enable a tractable optimization formulation.

\subsection{Vehicle Dynamics}

Different vehicle models are adopted in the literature depending on the required fidelity \cite{rajamani2011vehicle}. Here, a simplified dynamic bicycle model consistent with \cite{rajamani2011vehicle, zhu2025Diff} is used. The steering angle is assumed to be small, lateral tire forces are linear in the slip angles, and longitudinal slip is neglected. These assumptions are reasonable for highway and mild urban driving, where curvature and longitudinal transients remain moderate.

With small-slip approximations, the slip angles become
\begin{equation}
    \alpha_f \approx \delta - \frac{v_y + l_f\omega}{v_x},
    \qquad
    \alpha_r \approx -\,\frac{v_y - l_r\omega}{v_x},
\end{equation}
and in the linear tire region $F_{y,f}=C_f\alpha_f$, $F_{y,r}=C_r\alpha_r$.

In such conditions, let  $(p_x,p_y)$ denote the center-of-mass position, $\psi$ the yaw angle, $(v_x,v_y)$ the body-frame velocities, $\omega$ the yaw rate, and $a$ and $\delta$ the longitudinal acceleration and steering angle, then the vehicle dynamics model is given by:
\begin{equation}\label{vehicle-dynamics}
\begin{cases}
    \dot{p}_x &= v_x \cos\psi - v_y \sin\psi \\
    \dot{p}_y &= v_x \sin\psi + v_y \cos\psi \\
    \dot{\psi} &= \omega \\
    \dot{v}_x &= a + \omega v_y \\
    \dot{v}_y &= -\omega v_x + \tfrac{1}{m}(F_{y,f}+F_{y,r}) \\
    \dot{\omega} &= \tfrac{1}{I_z}(F_{y,f}l_f - F_{y,r}l_r).
\end{cases}
\end{equation}

Under the stated assumptions, the system can be approximated as differentially flat with respect to the position coordinates $(p_x, p_y)$ in the sense that the remaining states and inputs can be expressed as functions of these outputs and a finite number of their derivatives, up to modeling approximations \cite{Rathinam1995DifferentialFO}. This justifies planning directly in the geometric space while maintaining dynamic consistency.

Let
\begin{equation}
    \boldsymbol{v}_I =
    \begin{bmatrix}
        \dot{p}_x \\ \dot{p}_y
    \end{bmatrix},
    \qquad
    v = \|\boldsymbol{v}_I\|,
    \qquad
    \theta = \tan^{-1}\!\left(\frac{\dot{p}_y}{\dot{p}_x}\right),
\end{equation}
and define the side-slip angle $\beta = \tan^{-1}(v_y/v_x)$. Since $\theta=\psi+\beta$ and $v_x=v\cos\beta$, $v_y=v\sin\beta$, small-slip conditions ($|\beta|\ll1$) imply $v_x\gg v_y$ and $\dot{v}_y \approx \dot{v}\beta + v\dot{\beta}$.

Substituting into the lateral dynamics and linear tire model yields the approximate side-slip dynamics
\begin{equation}
v\dot{\beta} + \dot{v}\beta
=
-\frac{C_f+C_r}{m}\beta
-\omega v
+\frac{C_f}{m}\delta
+\frac{C_r l_r - C_f l_f}{mv}\omega.
\end{equation}

For typical passenger vehicle parameters, the lateral dynamics, governed by the cornering stiffness coefficients $C_f$ and $C_r$, induce a characteristic time constant on the order of $m/(C_f + C_r)$, which is typically in the range of tens of milliseconds. This is significantly smaller than the time scales associated with longitudinal motion and path evolution in motion planning problems (on the order of seconds). This separation of time scales motivates the quasi–steady-state approximation, in which the transient dynamics of $\beta$ are neglected so that $v\dot{\beta} + \dot{v}\beta \approx 0$, which allows the side-slip angle to be approximated algebraically as a function of the steering input, curvature, and velocity as follows:

\begin{equation}
\beta \approx
\frac{C_f}{C_f + C_r}\,\delta
+
\frac{1}{C_f + C_r}
\left(
\frac{C_r l_r - C_f l_f}{v}
- m v
\right)\omega.
\end{equation}

Differentiating $\theta=\psi+\beta$ gives $\dot{\theta}=\omega+\dot{\beta}$. Under the time-scale separation assumption, $\dot{\beta}$ is small relative to $\omega$, yielding the approximation $\omega \approx \dot{\theta}$. Considering the definition of curvature:
\begin{equation}
k =
\frac{\dot{p}_x \ddot{p}_y - \dot{p}_y \ddot{p}_x}
{(\dot{p}_x^2 + \dot{p}_y^2)^{3/2}},
\end{equation}
it follows that
\begin{equation}
\omega \approx v k.
\end{equation}

Thus $\beta$ and $\omega$ can be expressed approximately in terms of $v$, $k$, and $\delta$, all of which depend on derivatives of $(p_x,p_y)$. The remaining states follow from
\begin{equation}
    \psi = \theta - \beta,
    \qquad
    v_x \approx v,
    \qquad
    v_y \approx v\beta,
\end{equation}
and the longitudinal input from
\begin{equation}
    a \approx \dot{v} - v^2 k \beta.
\end{equation}

Under the small-slip and linear-tire assumptions, the relations above establish an \emph{approximate} differential flatness property of the dynamic bicycle model with flat outputs $(p_x,p_y)$. In particular, the quasi–steady-state approximation $v\dot{\beta} + \dot{v}\beta \approx 0$ assumes that the side-slip dynamics evolve on a faster time scale than the longitudinal and yaw dynamics, allowing $\beta$, $\omega$, and the corresponding inputs to be recovered algebraically from derivatives of the position trajectory.

This approximation is not exact and is introduced to obtain a tractable relation between the flat outputs and the remaining states and inputs. While it enables reconstruction of dynamically consistent trajectories, it introduces modeling error in quantities such as $\beta$ and $\delta$. These deviations affect reconstruction fidelity but not the geometric feasibility of the trajectory, which remains defined by $(p_x(t),p_y(t))$ and its derivatives.

\subsection{Bézier Curves}

Bézier curves are widely used in motion planning algorithms that rely on parametric trajectory representations due to their favorable geometric and analytical properties \cite{zayou2025Graph, marcucci2023motion}. In particular, Bézier curves provide a compact polynomial parameterization of trajectories in terms of a finite set of control points, which makes them especially suitable for optimization-based formulations. Formally, a Bézier curve is a polynomial mapping $\boldsymbol{r} : [0,1] \rightarrow \mathbb{R}^n$ defined as follows: 

\begin{definition}[Bézier Curve]
A Bézier curve $\boldsymbol{r}(s)$ of degree $m \in \mathbb{N}$ is defined as
\begin{equation}
    \boldsymbol{r}(s) \coloneqq \sum_{i=0}^{m} b_{i,m}(s) P_i,
\end{equation}
where $b_{i,m}(s)$ are the Bernstein polynomials of degree $m$.
\end{definition}

Several properties make Bézier curves attractive for motion planning. First, they interpolate their first and last control points,
\begin{equation}
    \boldsymbol{r}(0) = P_0,
    \qquad
    \boldsymbol{r}(1) = P_m,
\end{equation}
which allows boundary conditions to be imposed directly.

Second, the derivative of a Bézier curve of degree $m$ is a Bézier curve of degree $m-1$, so velocity and acceleration constraints can be expressed as linearly in the control points.

Most importantly, every point $\boldsymbol{r}(s)$ lies in the convex hull of its control points, since the Bernstein polynomials form a nonnegative partition of unity. Thus, constraining the control points to a convex set guaranties that the entire curve remains within that set.

However, in motion planning problems involving kinematic and dynamic constraints, it is necessary to relate the curve parameter $s \in [0,1]$ to physical time, as the parameter $s$ describes solely geometric progression along the curve. To this end, an auxiliary function is introduced that maps the curve parameter to time.

Let $t = h(s)$ be a strictly increasing polynomial function defined on $[0,1]$, commonly referred to as a \emph{time-scaling polynomial}. Monotonicity ensures invertibility, and its inverse is denoted by $s = g(t)$, where $g : [0,T] \rightarrow [0,1]$. Then, the time-parameterized trajectory is obtained by composition,
\begin{equation}
    \boldsymbol{q}(t) = \boldsymbol{r}(g(t)).
\end{equation}

\subsection{Optimization over Graphs of Convex Sets}

Using the trajectory representation introduced previously, the motion planning problem can be formulated directly in terms of the time-parameterized position trajectory $\boldsymbol{q}(t)=\boldsymbol{r}(g(t))$. Under the previously stated assumptions regarding the operating conditions of the vehicle, planning is carried out in the output space without explicitly introducing control inputs. The problem can therefore be written as
\begin{align}
    \min_{y(\cdot)} \quad & \int_0^T \ell(y(t))\,dt \\
    \text{s.t.} \quad 
    & y(0)=y_0, \quad y(T)\in Y_{\mathrm{goal}}, \\
    & y(t)\in Y_a, \\
    & \dot y(t)\in Y_{\mathrm{diff}},
\end{align}
where $y(t)=\boldsymbol{q}(t)$ and $Y_{\mathrm{diff}}$ encodes algebraic constraints on trajectory derivatives. While inputs are eliminated, the geometric constraint $y(t)\in Y_a$ remains nonconvex.

Following \cite{marcucci2023motion}, the free region is approximated as a finite union of convex polytopes,
\begin{equation}
    Y_a \approx \bigcup_{v\in\mathcal{V}} \mathcal{C}_v,
\end{equation}
and represented by a directed graph $\mathcal{G}=(\mathcal{V},\mathcal{E})$, where each vertex corresponds to a region $\mathcal{C}_v$ and each edge encodes an admissible transition. If the trajectory within region $\mathcal{C}_v$ is represented by a Bézier curve with control points $P_{k,v}$, enforcing $P_{k,v}\in\mathcal{C}_v$ guarantees containment of the entire segment. 

To capture connectivity, binary variables $y_{ij}\in\{0,1\}$ are introduced, indicating whether the edge $(i,j)\in\mathcal{E}$ is selected to compose the path. The path selection is encoded through flow conservation:
\begin{equation}
    \sum_{j:(i,j)\in\mathcal{E}} y_{ij}
    -
    \sum_{k:(k,i)\in\mathcal{E}} y_{ki}
    = 0,
\end{equation}
with appropriate boundary conditions at source and target vertices.

In this formulation, constraints of the form
\begin{equation}
    A  y_{ij}P_{k,i} \le b\, y_{ij},
    \qquad
    y_{ij} P_{m,i} = y_{ij} P_{0,j},
\end{equation}
are enforced only when $y_{ij}=1$. The constraints above introduce bilinear terms involving products of binary and continuous variables. To recover a convex formulation, a standard lifting technique is employed, introducing auxiliary variables that represent these products and allow the constraints to be expressed linearly in the lifted space.  Define
\begin{equation}
    Z_{ij,k,i} = y_{ij} P_{k,i},
\end{equation}
which eliminates explicit products and yields
\begin{equation}
    A Z_{ij,k,i} \le b\, y_{ij},
    \qquad
    Z_{ij,m,i}=Z_{ij,0,j}.
\end{equation}
When $y_{ij}=0$, the lifted variables vanish; when $y_{ij}=1$, the original trajectory variables are recovered.

Let $s,t\in\mathcal{V}$ denote the source and target vertices. The lifted formulation becomes the mixed-integer convex program
\begin{align}
    \min \quad 
    & \sum_{(i,j)\in\mathcal{E}} \ell_{ij}(Z_{ij,k,i}) \\
    \text{s.t.} \quad 
    & A Z_{ij,k,i} \le b\, y_{ij}, \\
    & A_{\mathrm{diff}} D Z_{ij,k,i}
      \le
      b_{\mathrm{diff}}\, y_{ij}, \\
    & \sum_{j:(i,j)\in\mathcal{E}} y_{ij}
      -
      \sum_{k:(k,i)\in\mathcal{E}} y_{ki}
      = 0, \\
    & y_{ij}\in\{0,1\},
\end{align}
where $D$ is a differentiation operator for the Bézier curve.

Relaxing the binary constraints to $y_{ij}\in [0, 1]$ yields a convex program that provides a lower bound on the original problem. The resulting solution can be interpreted as a fractional flow over the graph, which does not correspond to a single feasible path but can be used to guide the selection of a discrete solution through rounding procedures. It should be noted that the rounding step is not guaranteed to preserve optimality or feasibility with respect to the relaxed solution, and its effectiveness depends on the structure of the graph and the tightness of the relaxation.

\subsection{Nonlinear Discrete-Time Optimal Control}

To provide a baseline for comparison, the motion planning problem is also formulated and solved as a nonlinear discrete-time optimal control problem. In contrast to the proposed \gls{GCS} formulation, the \gls{NLP} retains the full nonlinear vehicle dynamics and enforces obstacle avoidance and dynamic constraints directly in the original state space.

Let the state vector $\boldsymbol{x}$ and the control input $\boldsymbol{u}$ be defined as
\begin{equation}
    \boldsymbol{x}
    =
    \begin{bmatrix}
        p_x & p_y & \psi & v_x & v_y & \omega
    \end{bmatrix}^\top,     \boldsymbol{u}
    =
    \begin{bmatrix}
        a & \delta
    \end{bmatrix}^\top,
\end{equation}

along with the continuous-time dynamic equation:
\begin{equation}
    \dot{\boldsymbol{x}} = f(\boldsymbol{x},\boldsymbol{u}),
\end{equation}

For numerical optimization, the system is discretized over a finite horizon of length $T$ using a fixed sampling time $\Delta t$, with $N = T/\Delta t$ steps. A forward Euler discretization is adopted for simplicity. While higher-order integration schemes could improve accuracy, the chosen discretization is sufficient for the comparative purposes of this study. The discretization results in the following discrete-time model: 
\begin{equation}
    \boldsymbol{x}_{k+1}
    =
    \boldsymbol{x}_k
    +
    \Delta t\, f(\boldsymbol{x}_k,\boldsymbol{u}_k),
    \qquad
    k = 0,\dots,N-1.
\end{equation}
The optimization variables are, therefore, the sequences $\{\boldsymbol{x}_k\}_{k=0}^{N}$ and $\{\boldsymbol{u}_k\}_{k=0}^{N-1}$. The baseline problem is formulated as the \gls{NLP}:
\begin{equation}
\begin{aligned}
    \min_{\{\boldsymbol{x}_k,\boldsymbol{u}_k\}}
    \quad
    & \sum_{k=0}^{N-1}
      \ell(\boldsymbol{x}_k,\boldsymbol{u}_k)
      +
      \ell_f(\boldsymbol{x}_N) \\
    \text{s.t.} \quad
    & \boldsymbol{x}_{k+1}
      =
      \boldsymbol{x}_k
      +
      \Delta t\, f(\boldsymbol{x}_k,\boldsymbol{u}_k), \\
    & \boldsymbol{x}_0 = \boldsymbol{x}_{\mathrm{init}}, \quad \boldsymbol{x}_N \in \mathcal{X}_{\mathrm{goal}}, \\
    & \boldsymbol{x}_k \in \mathcal{X}_{\mathrm{adm}}, \\
    & \boldsymbol{u}_k \in \mathcal{U}_{\mathrm{adm}},
    \qquad
    k = 0,\dots,N-1.
\end{aligned}
\end{equation}

The stage cost penalizes control effort and aggressive maneuvers and is chosen as
\begin{equation}
    \ell(\boldsymbol{x}_k,\boldsymbol{u}_k)
    =
    w_a a_k^2
    +
    w_\delta \delta_k^2
    +
    w_{\dot{\delta}} (\delta_k - \delta_{k-1})^2,
\end{equation}
where $w_a$, $w_\delta$, and $w_{\dot{\delta}}$ are positive weights. This structure penalizes large longitudinal accelerations, large steering angles, and rapid steering variations, promoting smoothness and passenger comfort. A terminal cost $\ell_f(\boldsymbol{x}_N)$ penalizes deviation from the goal configuration. It should be noted that the objective functions in the GCS and NLP formulations are not identical. While the NLP directly penalizes control inputs and their variations, the GCS formulation penalizes higher-order derivatives of the trajectory as convex surrogates for these quantities. Therefore, the comparison focuses on the resulting trajectory properties rather than exact optimality with respect to a common cost function.

Kinematic feasibility is enforced through the constraints: 

\begin{equation}
    v_{\min} \leq v_{x,k}\le v_{\max}, \quad |\delta_k| \le \delta_{\max}
\end{equation}

Obstacle avoidance is enforced in configuration space by approximating the ego vehicle with two circles aligned with its longitudinal axis and representing each obstacle as an ellipse. Using the Minkowski sum, denoted by $\oplus$, each obstacle is inflated by the ego geometry. Hence, the obstacle avoidance constraints can be formulated as:
\begin{equation}
    d\big((p_{x,k},p_{y,k}), \mathcal{O}_i \oplus \mathcal{B}\big) \ge 0, \quad k = 0,1,..., N \quad i = 1,2... ,
\end{equation}

Resulting in constraints that are nonlinear and nonconvex in the state variables.

The \gls{NLP} is solved using a general-purpose solver, optimizing the full state and control trajectories simultaneously. It serves as a baseline representing a direct transcription of the nonlinear optimal control problem against which the approximate \gls{GCS}-based formulation is compared under consistent modeling assumptions.

\section{Problem Formulation}

This section formalizes the motion planning problem for autonomous vehicles as a finite-dimensional mixed-integer convex optimization problem. The formulation integrates geometric path parameterization, polynomial time scaling, and a graph-of-convex-sets representation of the free space.

The admissible space is modeled as a finite union of convex polytopes,
\begin{equation}
Y_a = \bigcup_{v \in \mathcal{V}} \mathcal{C}_v,
\end{equation}
where each $\mathcal{C}_v \subset \mathbb{R}^2$ is a convex set associated with a vertex $v$ of the graph of convex sets $\mathcal{G} = (\mathcal{V}, \mathcal{E})$. 

Two families of control points are attached to each vertex:
\begin{equation}
\{P_{0,v}, \dots, P_{m,v}\} \subset \mathbb{R}^2,
\qquad
\{\tau_{0,v}, \dots, \tau_{m,v}\} \subset \mathbb{R}.
\end{equation}

The control points $P_{l,v}$ parameterize the geometric Bézier segment associated with region $\mathcal{C}_v$, while $\tau_{l,v}$ parameterize the corresponding time-scaling polynomial.

The motion planning problem over the \gls{GCS} framework is therefore formulated by imposing constraints and objective terms directly on the polynomial coefficients attached to each vertex. These constraints encode geometric feasibility, dynamic feasibility, and temporal consistency, thereby transforming the original infinite-dimensional optimal control problem into a finite-dimensional mixed-integer convex program, as detailed in the subsequent sections.

\subsection{Representations for velocity and acceleration}

The geometric path and the time-scaling are represented by two separate Bézier curves, 
$\boldsymbol{r} = \boldsymbol{r}(s)$ and $t = h(s)$, which can be composed to produce a time-domain representation of the trajectory $\boldsymbol{q}(t)$. Since $h$ is assumed to be strictly increasing, it is invertible and admits an inverse $s = h^{-1}(t) = g(t)$. The trajectory can therefore be written equivalently as
\begin{equation}
    \boldsymbol{q}(h(s)) = \boldsymbol{r}(s),
    \qquad
    \text{and}
    \qquad
    \boldsymbol{q}(t) = \boldsymbol{r}(g(t)).
\end{equation}

Within this formulation, the derivative of $\boldsymbol{q}(t)$ with respect to time, denoted $\dot{\boldsymbol{q}}(t)$, follows directly from the chain rule:
\begin{equation}
    \dot{\boldsymbol{q}}(t) 
    = 
    \frac{d}{dt} \boldsymbol{r}(g(t)) 
    = 
    \frac{d}{ds}\boldsymbol{r}(g(t)) \, \frac{d}{dt}g(t).
\end{equation}
Noting that $s = g(t)$ and denoting differentiation of a function $f(s)$ with respect to $s$ by $f'(s)$, this expression simplifies to
\begin{equation}
    \dot{\boldsymbol{q}}(t) 
    = 
    \boldsymbol{r}'(s)\,\dot{g}(t).
\end{equation}

A useful relation between $\dot{g}(t)$ and $h'(s)$ is obtained by differentiating the identity $t = h(s)$ with respect to $t$. Applying the chain rule yields
\begin{equation}
    1 = h'(s)\,\dot{g}(t),
\end{equation}

Substituting this relation into the expression for $\dot{\boldsymbol{q}}(t)$ gives the compact representation for the velocity vector of the curve
\begin{equation}
    \dot{\boldsymbol{q}}(t) 
    = 
    \frac{\boldsymbol{r}'(s)}{h'(s)}.
\end{equation}

Differentiating once more with respect to time yields the acceleration
\begin{equation}
    \ddot{\boldsymbol{q}}(t)
    =
    \frac{\boldsymbol{r}''(s) h'(s) - \boldsymbol{r}'(s) h''(s)}{h'(s)^3}.
\end{equation}

The acceleration vector $\ddot{\boldsymbol{q}}(t)$ can be decomposed into tangential and normal components relative to the geometric path. The unit tangent vector is defined as
\begin{equation}
    \boldsymbol{T}(s) = \frac{\boldsymbol{r}'(s)}{\|\boldsymbol{r}'(s)\|}.
\end{equation}

The tangential component of the acceleration is therefore
\begin{equation}
    \ddot{q}_T(t) 
    = 
    \boldsymbol{T}(s) \cdot \ddot{\boldsymbol{q}}(t)
    =
    \frac{\boldsymbol{T}(s)\cdot\boldsymbol{r}''(s)}{ h'(s)^2}
    -
    \frac{h''(s)}{h'(s)^3} \|\boldsymbol{r}'(s)\|.
\end{equation}

To characterize the normal component, consider the derivative of the unit tangent vector,
\begin{equation}
    \frac{d}{ds}\boldsymbol{T}(s)
    =
    \frac{\boldsymbol{r}''(s) - (\boldsymbol{T}(s)\cdot\boldsymbol{r}''(s))\,\boldsymbol{T}(s)}
         {\|\boldsymbol{r}'(s)\|}.
\end{equation}

The squared norm of this vector is
\begin{equation}
    \left\| \frac{d}{ds}\boldsymbol{T}(s) \right\|^2
    =
    \frac{\|\boldsymbol{r}''(s)\|^2 - (\boldsymbol{r}'(s)\cdot\boldsymbol{r}''(s))^2  \|\boldsymbol{r}'(s)\|^{-2}}
         {\|\boldsymbol{r}'(s)\|^2}.
\end{equation}

Using the identity
\begin{equation}
     \|\boldsymbol{r}'(s) \times \boldsymbol{r}''(s)\|^2 
     = 
     \|\boldsymbol{r}'(s)\|^2 \|\boldsymbol{r}''(s)\|^2 
     - 
     (\boldsymbol{r}'(s)\cdot\boldsymbol{r}''(s))^2,
\end{equation}
and the definition of curvature
\begin{equation}
    \kappa(s)
    =
    \frac{\|\boldsymbol{r}'(s) \times \boldsymbol{r}''(s)\|}
         {\|\boldsymbol{r}'(s)\|^3},
\end{equation}

the normal component of the acceleration can therefore be written as
\begin{equation}
    \ddot{q}_N(t)
    =
    \boldsymbol{\hat{N}}(s) \cdot \ddot{\boldsymbol{q}}(t)
    =
    \frac{\kappa(s)\|\boldsymbol{r}'(s)\|^2}{h'(s)^2}.
\end{equation}

where the unit normal vector $\boldsymbol{\hat{N}}(s)$ is defined as the normalized derivative of $\boldsymbol{T}(s)$.

The representations for velocity and acceleration derived above directly motivate the dynamic constraints introduced in the optimization problem. Moreover, when the angle $\beta$ is small, the tangential and normal components derived here provide approximations of the longitudinal and lateral accelerations of the vehicle.
 
\subsection{Constraints on Bézier Curves}

For each vertex $v \in \mathcal{V}$ and edge $(u,v)\in\mathcal{E}$ in the graph of convex sets $\mathcal{G} = (\mathcal{V}, \mathcal{E})$, constraints are imposed directly on the control points of the associated Bézier curves. These constraints encode temporal consistency, geometric feasibility, and dynamic plausibility.

\paragraph{Admissible space constraints}

The admissible space constraints follow directly from the convex-hull property of Bézier curves. Let the convex region associated with vertex $v \in \mathcal{V}$ be represented in half-space form as $\mathcal{C}_v = \{ x \in \mathbb{R}^2 \mid A_v x \le b_v \}$, where $A_v \in \mathbb{R}^{p_v \times 2}$ and $b_v \in \mathbb{R}^{p_v}$ define the supporting half-spaces of the polytope. Since a Bézier curve lies entirely within the convex hull of its control points, it suffices to impose
\begin{equation}
    A_v \boldsymbol{P}_{l,v} \le b_v,
    \qquad l = 0,\dots,m.
\end{equation}

\paragraph{Time-scaling plausibility}

To ensure that the time-scaling function remains physically meaningful, two conditions are imposed on its control points $\{\tau_{0,v},\dots,\tau_{m,v}\}$, which imply $\tau_{l,v} \ge 0$ for $l = 0,\dots,m.$

Second, the time-scaling must be strictly increasing to preserve invertibility of $t = h(s)$, which in terms of control points becomes:
\begin{equation}
    \tau_{l+1,v} - \tau_{l,v} \ge h'_{\min},
    \qquad l = 0,\dots,m-1,
\end{equation}

While $h'_{\min} = 0$ is sufficient for monotonicity, velocity and acceleration are rational functions of $h'(s)$, therefore, small values of $h'(s)$ amplify both velocity and acceleration.

\paragraph{Velocity constraints}

To ensure bounds on the magnitude of the velocity along the entire trajectory, note that the constraint $\|\dot{\boldsymbol{q}}(t)\| \le v_{\max}$ is equivalent to
\begin{equation}
    \|\boldsymbol{r}'(s)\|
    \le
    v_{\max} h'(s),
\end{equation}

This nonlinear constraint can be approximated by a polyhedral inner approximation of the unit ball. Let $a_k \in \mathbb{R}^2$, $k=1,\dots,F$, denote the outward normals of a polygon inscribed in the unit circle. The norm inequality can then be conservatively enforced through the set of linear constraints
\begin{equation}
    a_k^\top \boldsymbol{r}'(s)
    \le
    v_{\max} h'(s),
    \qquad
    k = 1,\dots,F.
\end{equation}

Since $\boldsymbol{r}'(s)$ and $h'(s)$ are Bézier curves of degree $m-1$, their control points are proportional to the forward differences of the original control points, namely $\boldsymbol{P}_{l+1,v}-\boldsymbol{P}_{l,v}$ and $\tau_{l+1,v}-\tau_{l,v}$, respectively. Using a polyhedral inner approximation of the unit circle with facet normals $a_k$, the constraint is enforced through linear inequalities of the form
\begin{equation}
    a_k^\top (\boldsymbol{P}_{l+1,v} - \boldsymbol{P}_{l,v})
    \le
    v_{\max} (\tau_{l+1,v} - \tau_{l,v}),
\end{equation}
for all relevant indices $l$ and $k$. This construction corresponds to an inner polyhedral approximation of the Euclidean unit ball, and therefore provides a conservative enforcement of the velocity constraint.

\paragraph{Continuity constraints}

To ensure smooth trajectories, continuity constraints are imposed on every edge $(u,v)\in\mathcal{E}$ of the graph of convex sets. In autonomous driving applications, continuity up to the third derivative is enforced for both the spatial curve $\boldsymbol{r}(s)$ and the time-scaling function $h(s)$.

Let the forward difference operators on the control points be defined as
\begin{equation}
\begin{aligned}
       \Delta^1 \boldsymbol{P}_{l,v} &= \boldsymbol{P}_{l+1,v} - \boldsymbol{P}_{l,v}, \\
       \Delta^2 \boldsymbol{P}_{l,v}
    &=
    \boldsymbol{P}_{l+2,v}
    - 2\boldsymbol{P}_{l+1,v}
    + \boldsymbol{P}_{l,v}, \\
    \Delta^3 \boldsymbol{P}_{l,v}
    &=
    \boldsymbol{P}_{l+3,v}
    - 3\boldsymbol{P}_{l+2,v}
    + 3\boldsymbol{P}_{l+1,v}
    - \boldsymbol{P}_{l,v}.
\end{aligned}
\end{equation}

Then continuity across edge $(u,v)$ is enforced by

\begin{equation}
\begin{aligned}
       \boldsymbol{P}_{m,u} &= \boldsymbol{P}_{0,v},, \\
       \Delta^1 \boldsymbol{P}_{m-1,u}
    &=
    \Delta^1 \boldsymbol{P}_{0,v}, \\
    \Delta^2 \boldsymbol{P}_{m-2,u}
    &=
    \Delta^2 \boldsymbol{P}_{0,v}, \\
    \Delta^3 \boldsymbol{P}_{m-3,u}
    &=
    \Delta^3 \boldsymbol{P}_{0,v}, \\
\end{aligned}
\end{equation}

The same construction is applied to the control points $\tau_{l,v}$.

These conditions enforce $C^3$ continuity across adjacent segments, ensuring that position, velocity, acceleration, and jerk remain continuous along the composed trajectory.

\subsection{Cost Function on Bézier Curves}

In autonomous driving, excessive accelerations degrade safety and passenger comfort. Although the tangential and normal acceleration expressions depend rationally on $h'(s)$ and do not admit direct convex bounds, their dominant components can be controlled through convex penalties on higher-order derivatives of the spatial and temporal Bézier curves.

In continuous form, the cost density can be written as
\begin{equation}
    \ell_v(s)
    =
    \alpha_1 \|\boldsymbol{r}''(s)\|
    +
    \alpha_2 \|\boldsymbol{r}'''(s)\|
    +
    \alpha_3 |h''(s)|
    +
    \alpha_4 |h'''(s)|,
\end{equation}
where $\alpha_i \ge 0$ are tuning parameters. The terms $\|r''(s)\|$ and $h''(s)$ act as convex surrogates for lateral and longitudinal acceleration components, respectively, and the third-order terms penalize acceleration variations, suppressing oscillatory behavior. In the finite-dimensional formulation, these quantities are expressed directly in terms of forward differences of the control points. 

The vertex-wise cost is then defined as the sum of convex penalties over all admissible indices:
\begin{equation}
\begin{aligned}
    J_v
    =
    \sum_{l=0}^{m-2}
    \alpha_1 \|\Delta^2 \boldsymbol{P}_{l,v}\|
    +
    \sum_{l=0}^{m-3}
    \alpha_2 \|\Delta^3 \boldsymbol{P}_{l,v}\| \\
    +
    \sum_{l=0}^{m-2}
    \alpha_3 |\Delta^2 \tau_{l,v}|
    +
    \sum_{l=0}^{m-3}
    \alpha_4 |\Delta^3 \tau_{l,v}|.
\end{aligned}
\end{equation}

The total objective of the optimization problem is obtained by summing $J_v$ over all active vertices selected by the graph flow variables.

This construction preserves computational tractability while promoting smooth, dynamically consistent trajectories but without explicit bounds on acceleration. A formulation that includes explicit bounds on acceleration, though, is of major interest in the context of vehicle dynamics and is an important research direction for future work.

\subsection{Dynamic obstacle avoidance}

Although the graph-of-convex-sets formulation assumes a static free space, dynamic obstacle avoidance can be incorporated when coarse predictions of obstacle motion are available. If an obstacle is expected to occupy a convex region $\mathcal{C}_v$ during an interval $[T_{\mathrm{in}},T_{\mathrm{out}}]$, collision avoidance can be enforced through temporal separation: the ego vehicle must either leave the region before $T_{\mathrm{in}}$ or enter it after $T_{\mathrm{out}}$. Since the entry and exit times correspond to the first and last control points of the time-scaling polynomial $h_v(s)$, these conditions are linear in the decision variables and remain compatible with the formulation.

In highway lane-following scenarios, where the ego and a leading vehicle share the same geometric path, collision avoidance reduces to regulating longitudinal timing. Heuristically, the traversal time of a segment is related to its arc length and average velocity. Based on this relation, constraints on entry and exit times can be used to regulate longitudinal separation from dynamic obstacles. Thus, bounding the time at which the ego exits a region implicitly bounds the average longitudinal speed, allowing safe separation from a leading vehicle without modifying the spatial curve $\boldsymbol{r}(s)$. It should be emphasized that this approach provides a heuristic approximation of dynamic obstacle avoidance and does not guarantee safety under arbitrary obstacle motion.

\section{Case Studies}

This section evaluates the proposed \gls{GCS} formulation in representative autonomous driving scenarios and compares the resulting trajectories with those obtained from a discrete-time optimal control formulation solved as a \gls{NLP} using IPOPT. The \gls{NLP} formulation is used as a baseline, as it closely approximates a direct solution of the nonlinear optimal control problem. The purpose of this comparison is not to claim that the \gls{GCS} approach is universally superior to existing methods, but rather to assess whether it can produce trajectories comparable to those obtained with a more exact formulation while offering a tractable alternative for motion planning. To mitigate the sensitivity of the \gls{NLP} formulation to local minima, the optimization was performed in successive stages, with each stage initialized from the solution of a simpler problem. This continuation strategy improves convergence robustness.

All scenarios were generated using the CommonRoad framework \cite{CommonRoad}, which provides standardized road geometries and obstacle representations and closely follows those considered in \cite{zhu2025Diff} to ensure comparability with established benchmarks. Although automated methods exist for decomposing free space into convex regions \cite{deits2015computing}, the convex sets used in this study were defined manually to isolate the performance of the proposed \gls{GCS} formulation from the decomposition procedure.

All experiments were conducted on a MacBook Pro equipped with an M2 Pro Max processor and 32 GB of RAM. Both formulations were implemented in Python using identical vehicle parameters, dynamic limits, and planning horizons ($\Delta t = 100$\,ms, $T = 10$\,s) to ensure a fair comparison. Execution time measurements over 500 runs, sufficient for convergence of statistics, are reported in Table~\ref{tab:execution time}.

\begin{table}[t]
\centering
\caption{Execution time measurements for NLP and GCS solutions in different scenarios}
\label{tab:execution time}
\begin{tabular}{|c | c | c|}
\hline
Scenario & NLP &  GCS \\
\hline
Static obstacle avoidance & 307.6ms $\pm$ 7.13ms & 12.1ms $\pm$ 1.04ms \\
\hline
Lane changing & 745.6ms $\pm$ 26.10ms & 13.9ms $\pm$ 2.43ms \\
\hline
Overtaking & 1179.9ms $\pm$ 32.77ms & 12.4ms $\pm$ 1.03ms \\
\hline
\end{tabular}
\end{table}

\subsection{Static Obstacle Avoidance}

\begin{figure*}
    \centering
     \subfloat[GCS solution.]{%
        \includegraphics[width=0.95\linewidth]{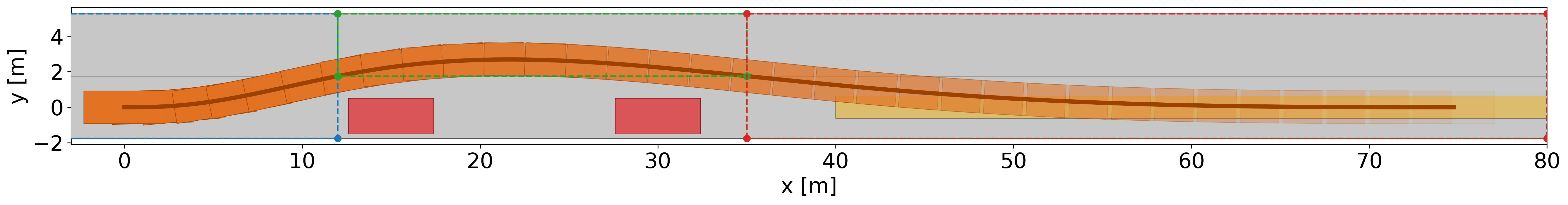}
        \label{fig:static-obstacles:gcs}
    }
     \hfill
     \subfloat[NLP solution.]{%
        \includegraphics[width=0.95\linewidth]{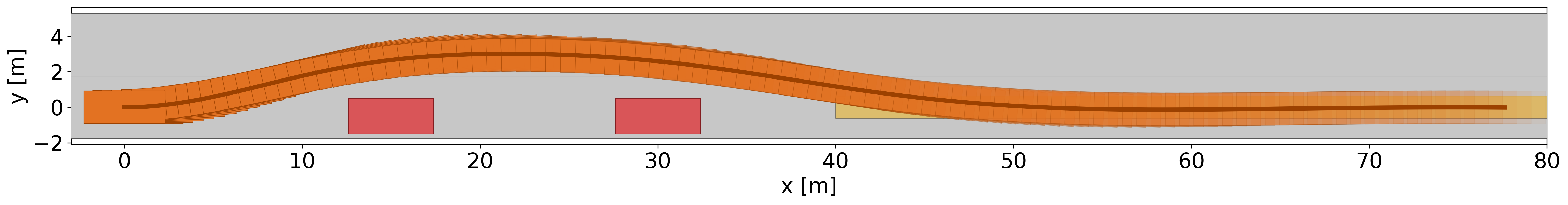}
        \label{fig:static-obstacles:nlp}
    }
    \caption{Trajectories obtained using (a) \gls{GCS} and (b) \gls{NLP} for the static obstacle avoidance scenario. The orange rectangles represent the occupancy of the ego vehicle; the red rectangles represent the occupancy of static obstacles; the yellow rectangle represents the target region. }
    \label{fig:static-obstacles}
\end{figure*}

The first scenario considers static obstacle avoidance on a structured two-lane roadway. Each lane has width $3.5\,\mathrm{m}$, and the two lanes span the interval $[-1.75\,\mathrm{m},\,5.25\,\mathrm{m}]$ along the $y$-axis. The ego vehicle and the obstacles are modeled with length $4.8\,\mathrm{m}$ and width $2.0\,\mathrm{m}$. Two static obstacles are placed at $(15\,\mathrm{m},-0.5\,\mathrm{m})$ and $(30\,\mathrm{m},-0.5\,\mathrm{m})$, respectively. 

The ego vehicle starts at $(0\,\mathrm{m},0\,\mathrm{m})$ with velocity $5\,\mathrm{m/s}$ and must reach the target region while achieving a final velocity of $8\,\mathrm{m/s}$. The environment consists of two lane-aligned corridors with a two obstacles.

Figure~\ref{fig:static-obstacles} presents the feasible and collision-free trajectories obtained using \gls{GCS} and \gls{NLP}, together with the convex free-space decomposition adopted in the \gls{GCS} formulation and the specification of the target region.

\begin{figure}
    \centering
    \subfloat[Minimum distance from obstacles.]{%
        \includegraphics[width=0.48\linewidth]{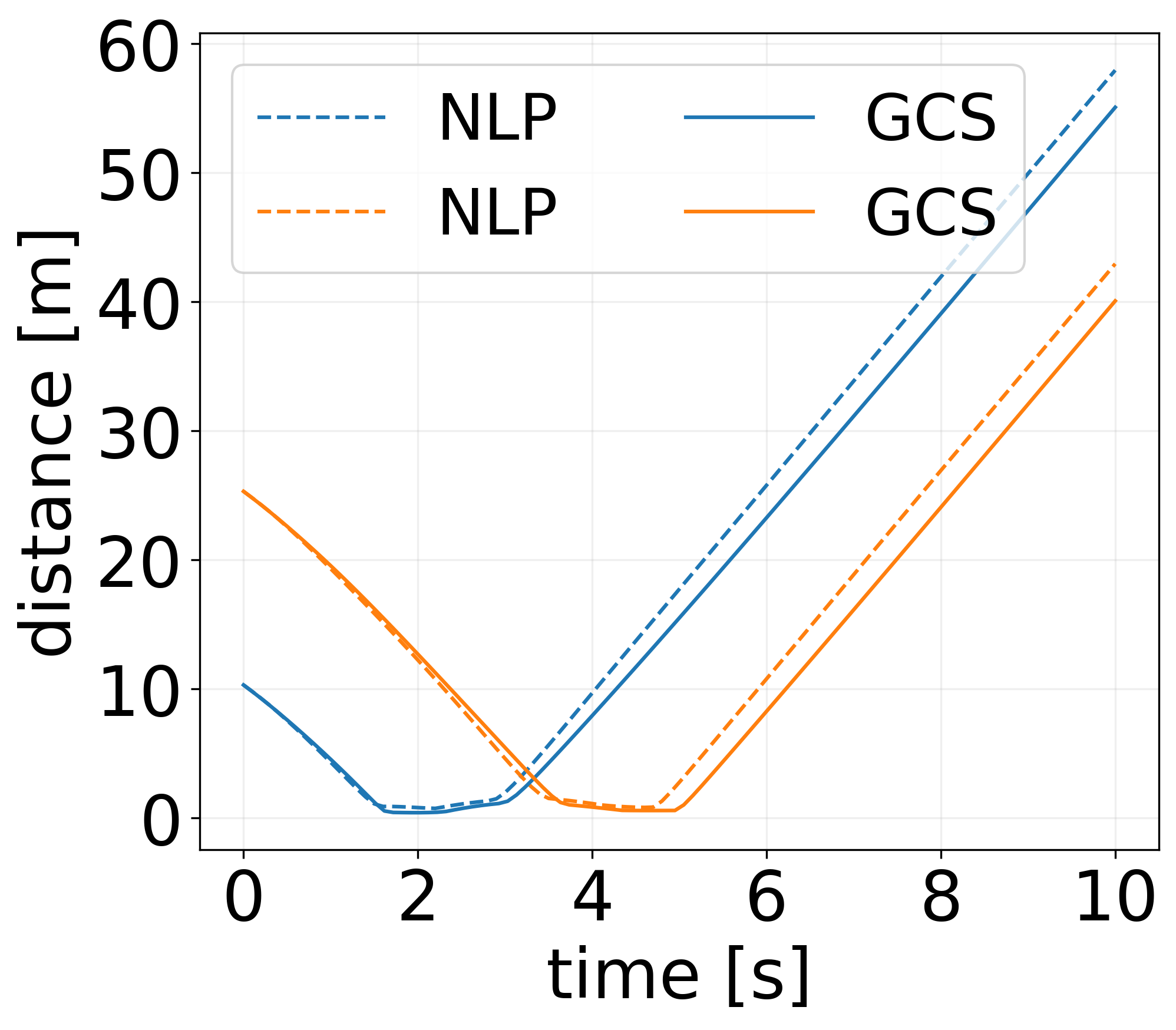}
        \label{fig:static-obstacles:distance}
    }
     \hfill
     \subfloat[Longitudinal velocity.]{%
        \includegraphics[width=0.48\linewidth]{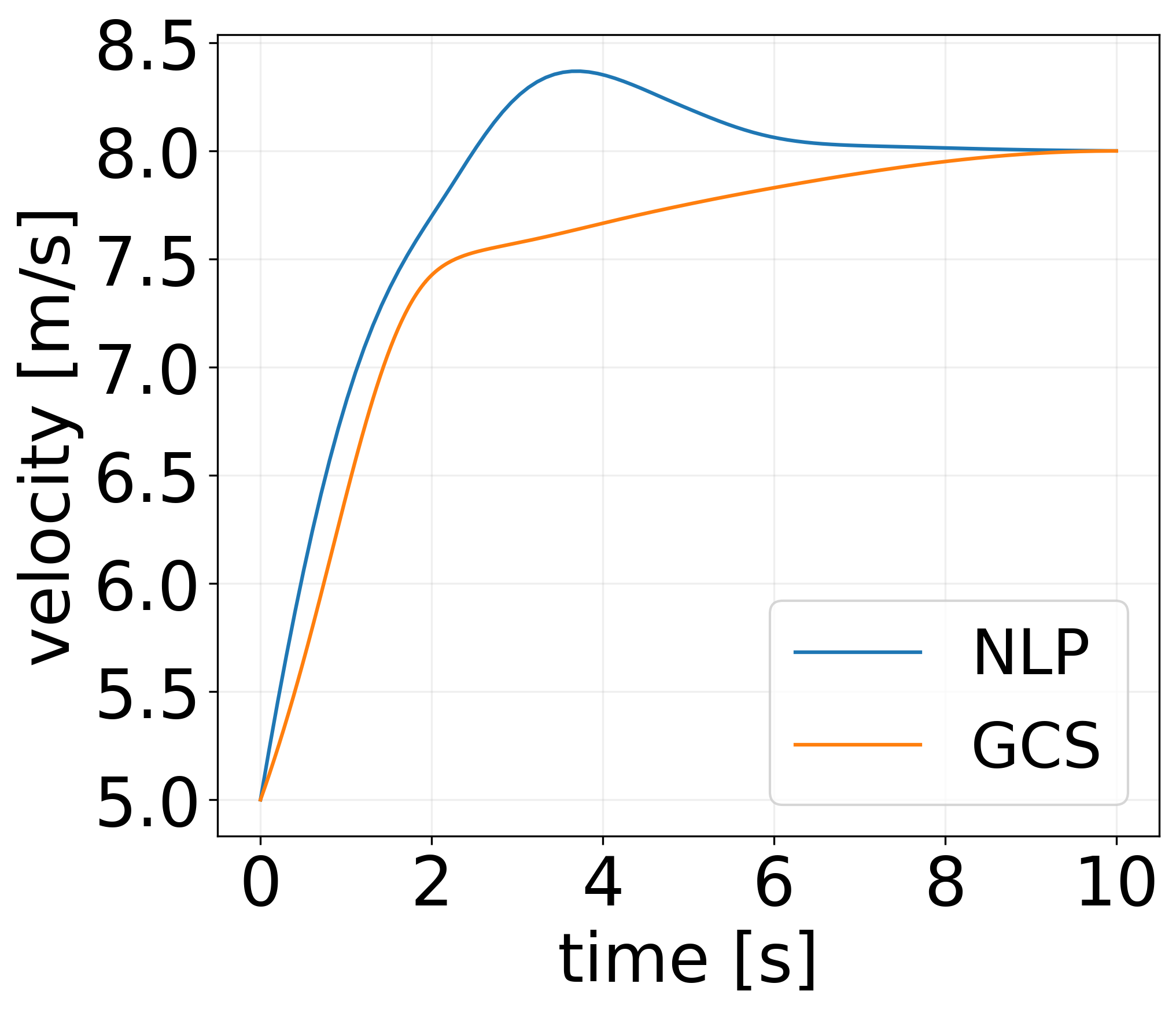}
        \label{fig:static-obstacles:velocity}
    }
     \hfill
     \subfloat[Longitudinal acceleration.]{%
        \includegraphics[width=0.48\linewidth]{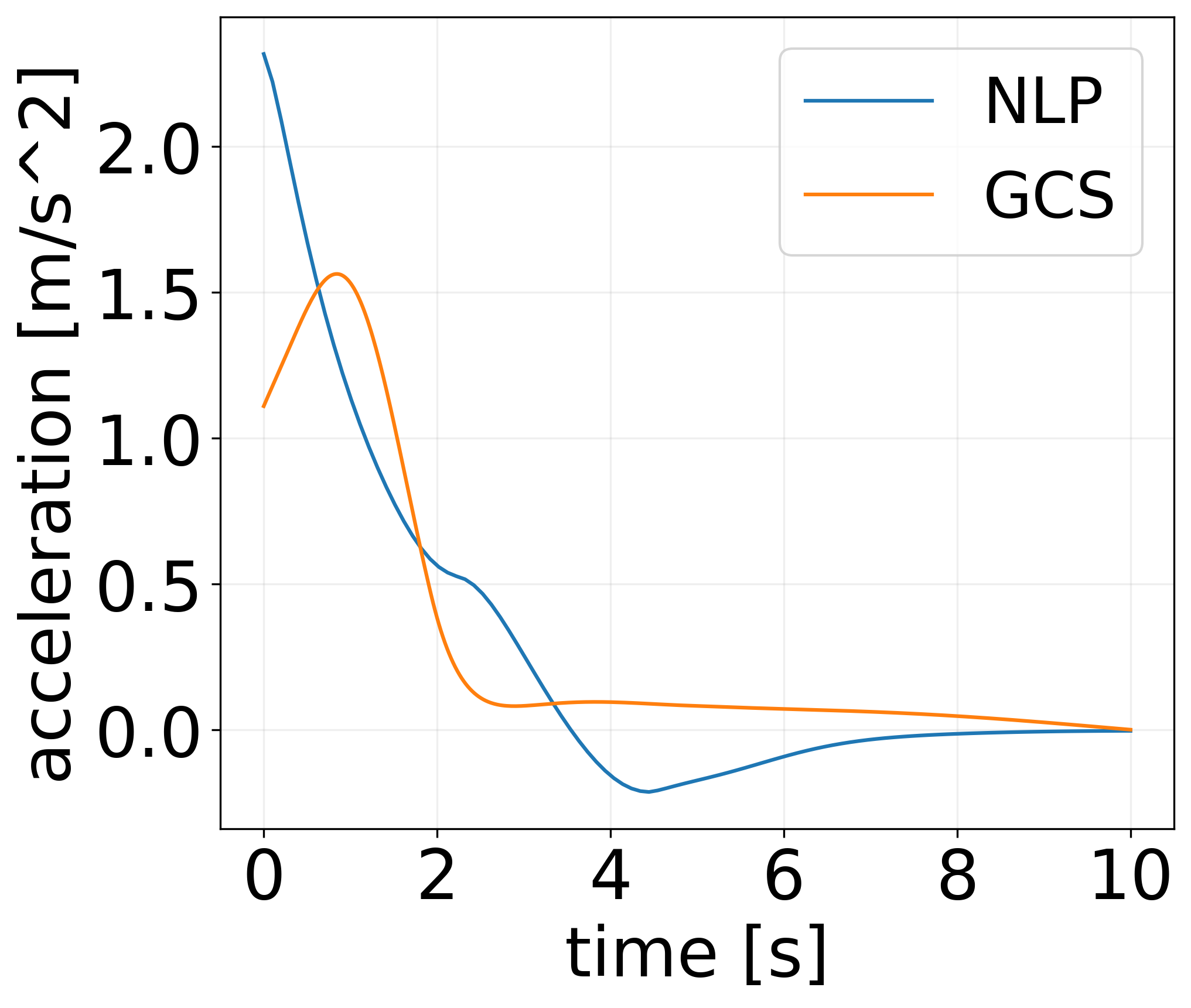}
        \label{fig:static-obstacles:acceleration}
    }
     \hfill
     \subfloat[Steering angle.]{%
        \includegraphics[width=0.48\linewidth]{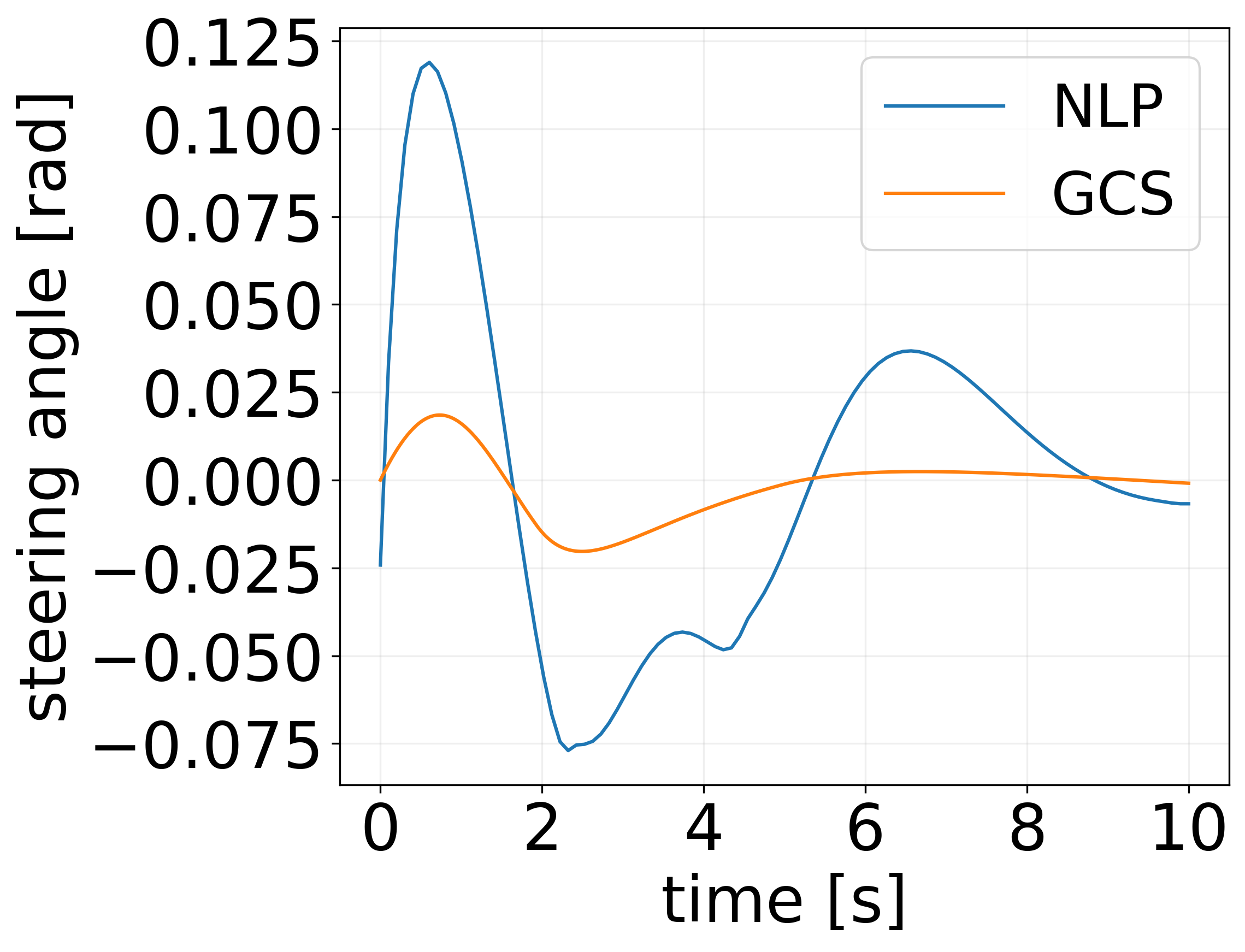}
        \label{fig:static-obstacles:steering}
    }
    \caption{(a) Minimum distance from obstacles, (b) longitudinal velocity, (c) longitudinal acceleration and (d) steering angle associated with the trajectories obtained using \gls{GCS} and \gls{NLP} for the static obstacle avoidance scenario.}
    \label{fig:static-obstacles:curves}
\end{figure}

Figure~\ref{fig:static-obstacles:curves} reports the minimum distance between the polygons representing the ego vehicle and the obstacles, as well as the resulting longitudinal velocity, longitudinal acceleration, and steering angle profiles. Both formulations respect velocity and dynamic limits, though the trajectories differ in steering angles and acceleration due to the distinct optimization structures.

\subsection{Lane-Changing}

\begin{figure*}
    \centering
     \subfloat[GCS solution.]{%
        \includegraphics[width=0.95\linewidth]{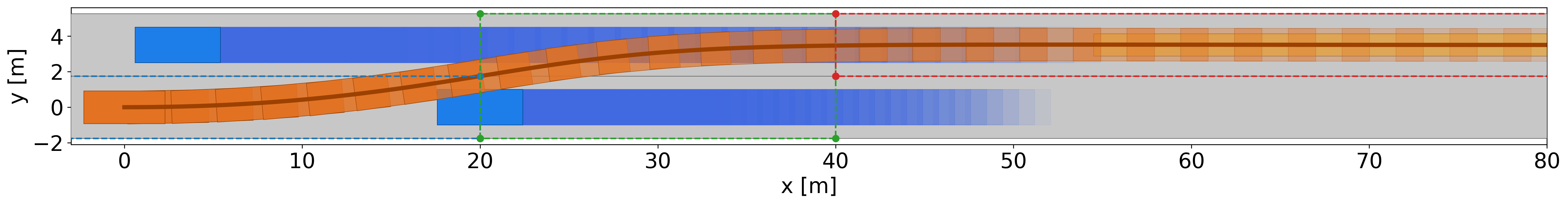}
        \label{fig:lane-change:gcs}
    }
     \hfill
     \subfloat[NLP solution]{%
        \includegraphics[width=0.95\linewidth]{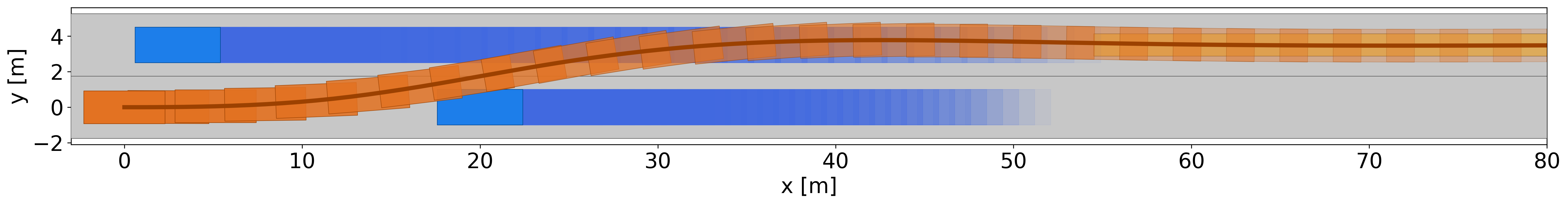}
        \label{fig:lane-change:nlp}
    }
    \caption{Trajectories obtained using (a) \gls{GCS} and (b) \gls{NLP} for the lane changing scenario. The orange rectangles represent the occupancy of the ego vehicle; the blue rectangles represent the occupancy of dynamic obstacles; the yellow rectangle represents the target region.}
    \label{fig:lane-change}
\end{figure*}

The second scenario evaluates a highway lane-changing maneuver in the presence of dynamic obstacles. The road geometry and structural assumptions follow those used in \cite{zhu2025Diff}. The first obstacle starts at $(20.0\,\mathrm{m}, 0.0\,\mathrm{m})$ with constant velocity $3\,\mathrm{m/s}$. The second obstacle starts at $(3.0\,\mathrm{m}, 3.5\,\mathrm{m})$ with constant velocity $5\,\mathrm{m/s}$. The ego vehicle begins at $(0\,\mathrm{m},0\,\mathrm{m})$ with velocity $8\,\mathrm{m/s}$ and must reach the target region with velocity $10\,\mathrm{m/s}$. The maneuver requires transitioning from the initial lane to the adjacent lane while satisfying dynamic constraints and avoiding collisions.

The selected free-space decomposition for the \gls{GCS} formulation, shown in Figure~\ref{fig:lane-change:gcs}, restricts lane transitions to an intermediate convex region (highlighted in green). Based on the predicted obstacle trajectories, a safe upper bound of $2.4\,\mathrm{s}$ is imposed on the time at which the ego vehicle must enter the intermediate region to pass ahead of the second obstacle without collision. The ego vehicle must then traverse this region within $4\,\mathrm{s}$ to ensure safe separation.

\begin{figure}
    \centering
    \subfloat[Minimum distance from obstacles.]{%
        \includegraphics[width=0.48\linewidth]{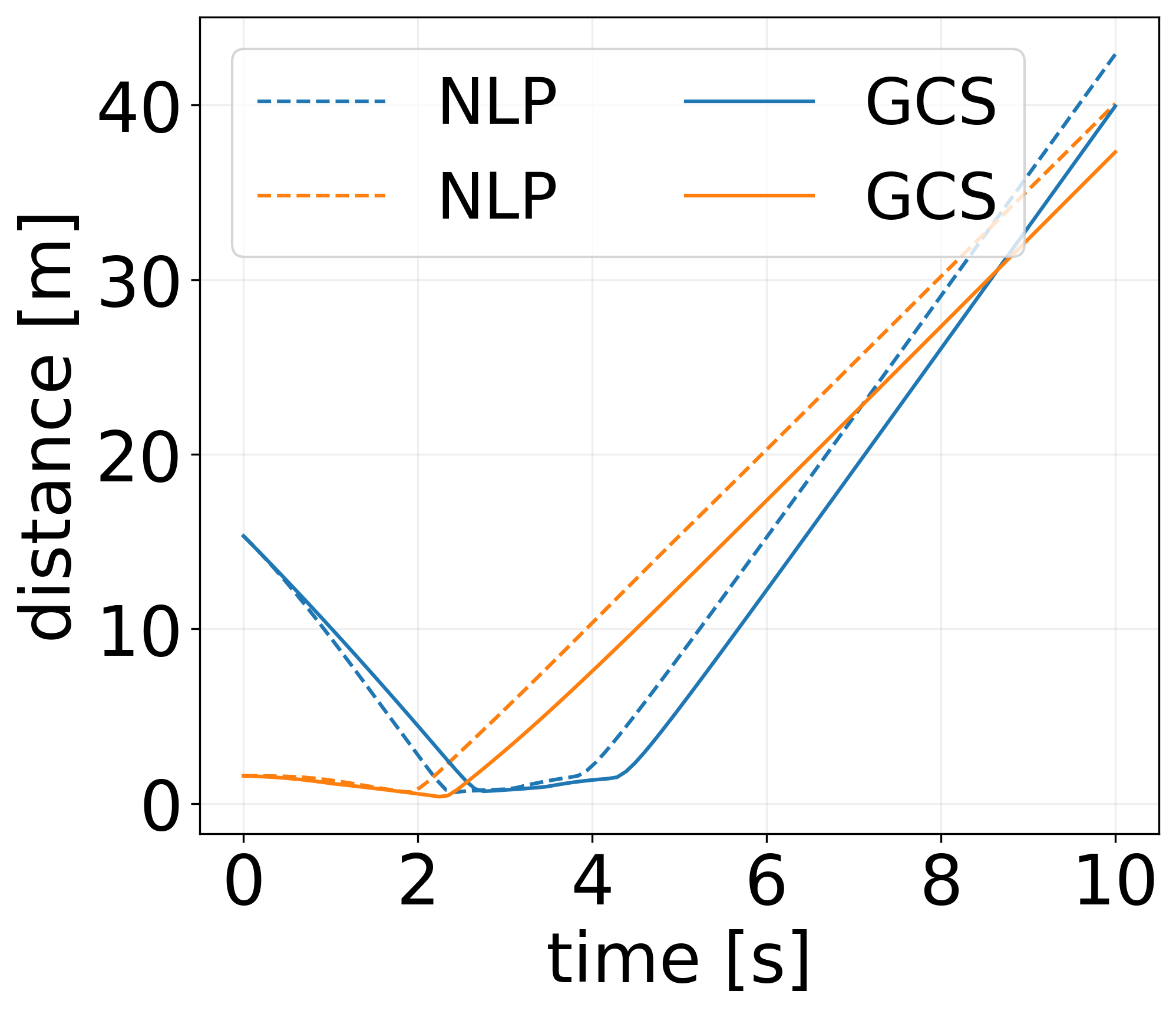}
        \label{fig:lane-change:distance}
    }
     \hfill
     \subfloat[Longitudinal velocity.]{%
        \includegraphics[width=0.48\linewidth]{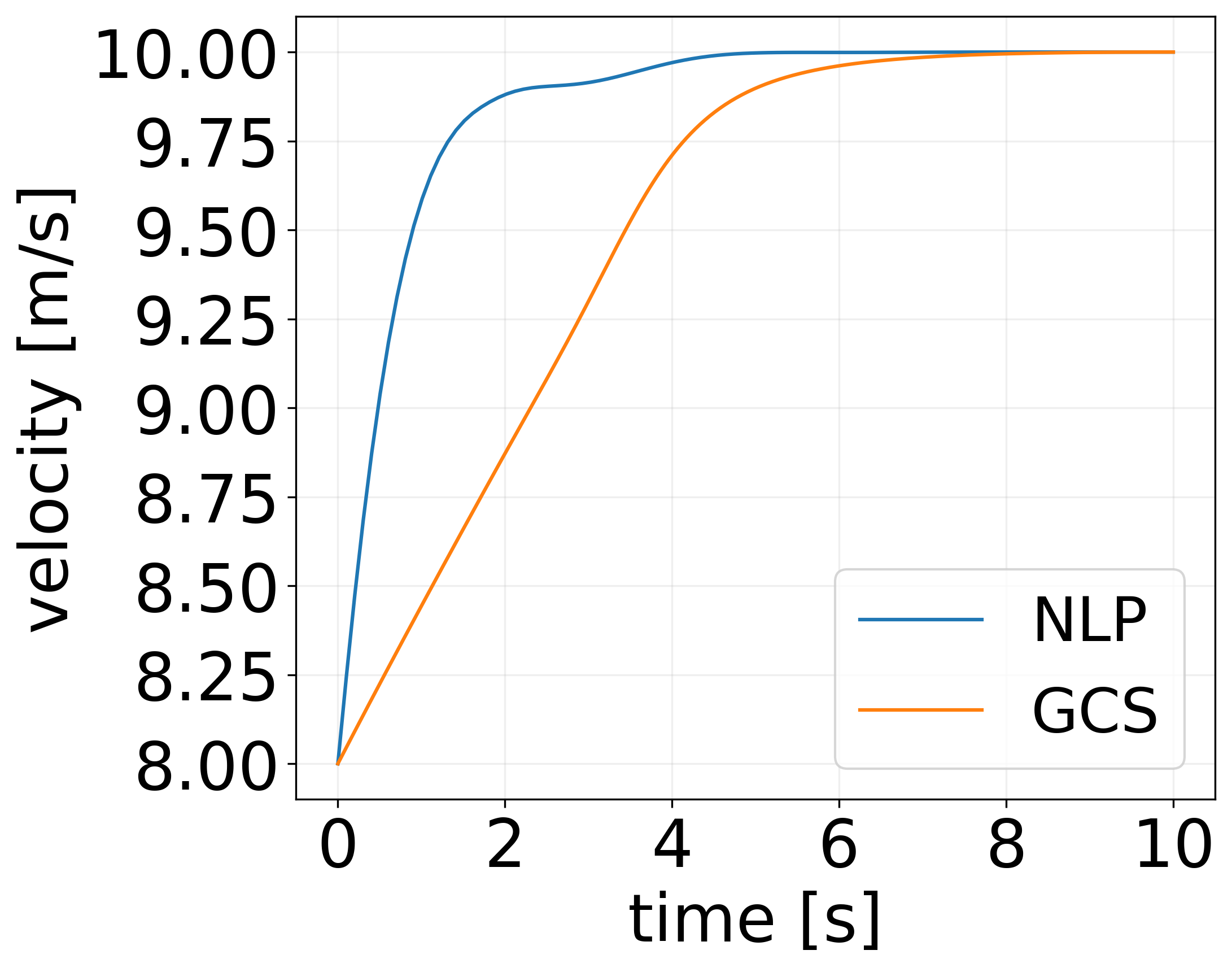}
        \label{fig:lane-change:velocity}
    }
     \hfill
     \subfloat[Longitudinal acceleration.]{%
        \includegraphics[width=0.48\linewidth]{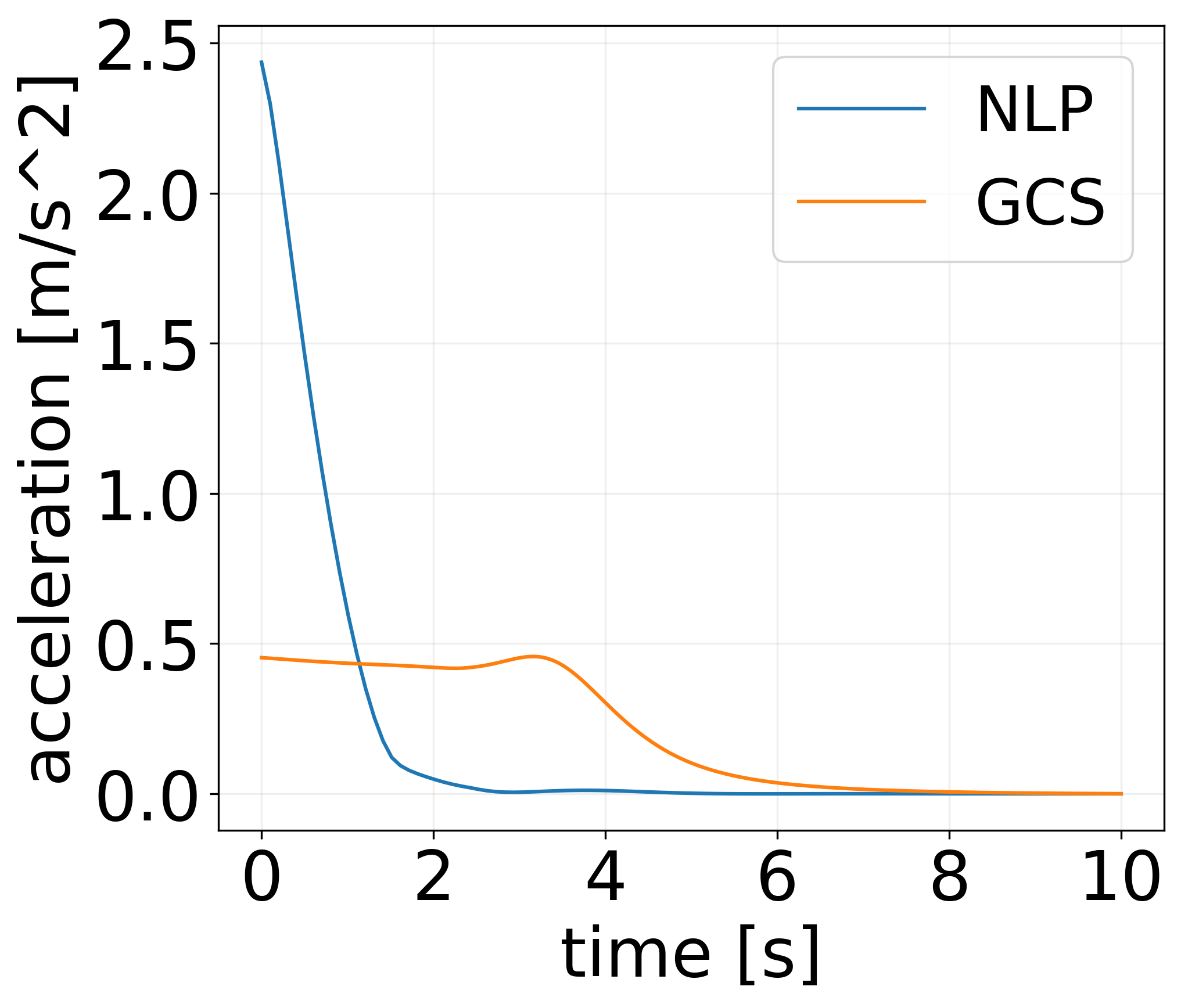}
        \label{fig:lane-change:acceleration}
    }
     \hfill
     \subfloat[Steering angle.]{%
        \includegraphics[width=0.48\linewidth]{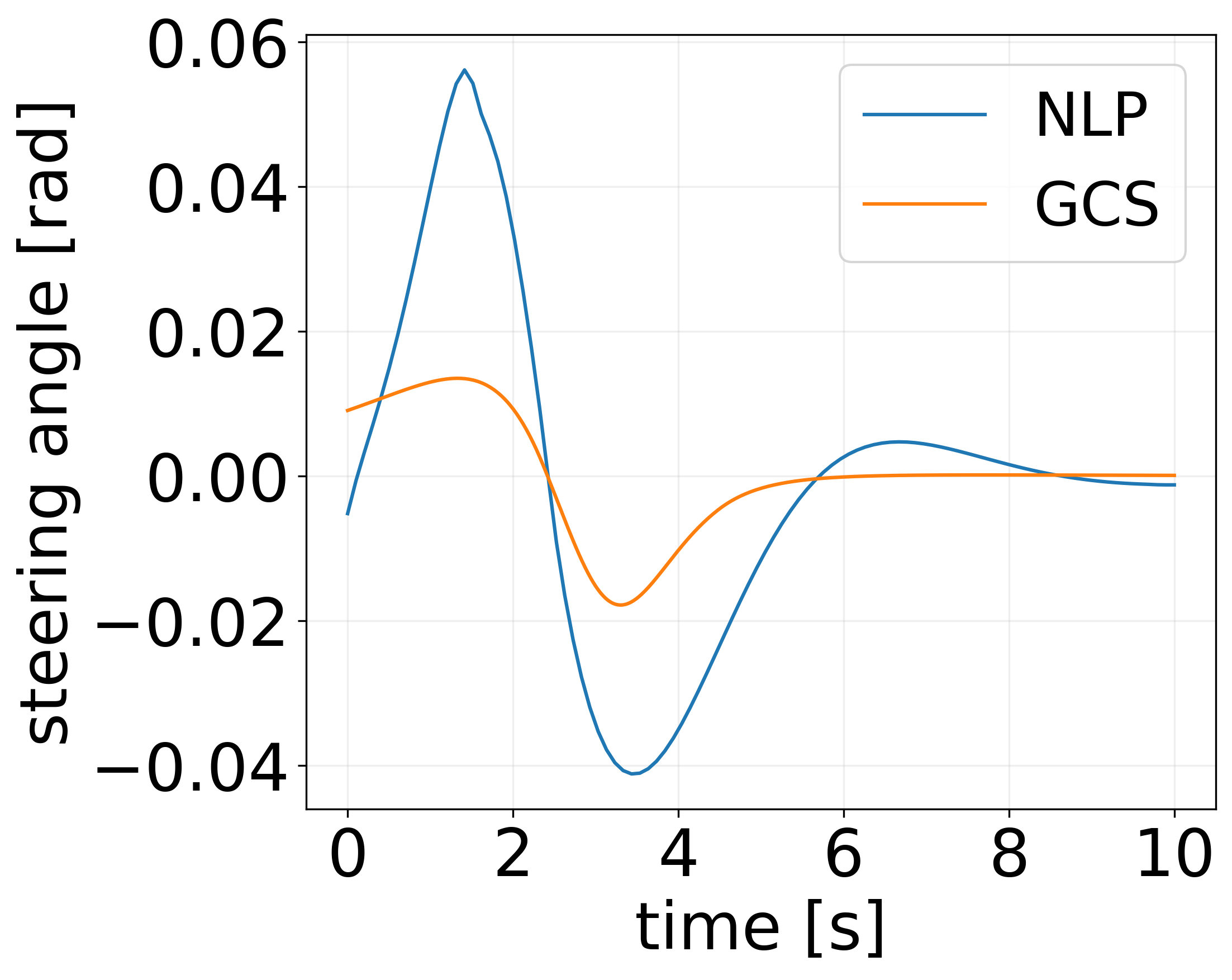}
        \label{fig:lane-change:steering}
    }
    \caption{(a) Minimum distance from obstacles, (b) longitudinal velocity, (c) longitudinal acceleration and (d) steering angle associated with the trajectories obtained using \gls{GCS} and \gls{NLP} for the lane changing scenario.}
    \label{fig:lane-change:curves}
\end{figure}

Figure~\ref{fig:lane-change} shows the resulting trajectories for both formulations. The associated minimum distance, longitudinal velocity, longitudinal acceleration, and steering angle profiles are presented in Figure~\ref{fig:lane-change:curves}. Both methods produce smooth and dynamically feasible lane-change trajectories, with differences arising primarily from the combinatorial structure of the \gls{GCS} formulation versus the local nature of the \gls{NLP} solution.

\subsection{Overtaking}

\begin{figure*}
    \centering
    \subfloat[GCS solution.]{%
        \includegraphics[width=0.95\linewidth]{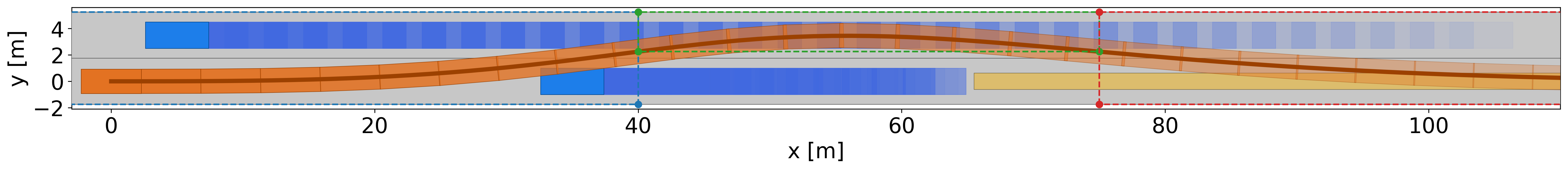}
        \label{fig:overtake:gcs}
    }
     \hfill
     \subfloat[NLP solution.]{%
        \includegraphics[width=0.95\linewidth]{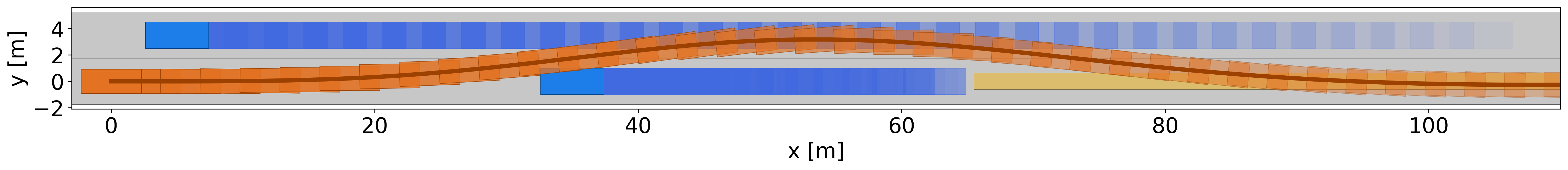}
        \label{fig:overtake:nlp}
    }
    \caption{Trajectories obtained using (a) \gls{GCS} and (b) \gls{NLP} for the overtaking scenario. The orange rectangles represent the occupancy of the ego vehicle; the blue rectangles represent the occupancy of dynamic obstacles; the yellow rectangle represents the target region.}
    \label{fig:overtake}
\end{figure*}

The third scenario considers an overtaking maneuver on a highway segment. The first obstacle starts at $(35.0\,\mathrm{m}, 0.0\,\mathrm{m})$ with velocity $3\,\mathrm{m/s}$, accelerates linearly to $8\,\mathrm{m/s}$, and subsequently decelerates back to $3\,\mathrm{m/s}$. The second obstacle starts at $(5.0\,\mathrm{m}, 3.5\,\mathrm{m})$ with constant velocity $10\,\mathrm{m/s}$.  The ego vehicle starts at $(0\,\mathrm{m},0\,\mathrm{m})$ with velocity $15\,\mathrm{m/s}$ and must reach the target region while maintaining $15\,\mathrm{m/s}$. The maneuver requires overtaking the leading vehicle while respecting dynamic limits and avoiding collisions with both obstacles.

The free-space decomposition adopted for the \gls{GCS} formulation, shown in Figure~\ref{fig:overtake:gcs}, again confines lane changes to an intermediate convex region. From the predicted obstacle trajectories, a safe upper bound of $2.7\,\mathrm{s}$ is derived for the instant at which the ego vehicle must enter this region to pass the leading vehicle while avoiding the second obstacle.

\begin{figure}
    \centering
    \subfloat[Minimum distance from obstacles.]{%
        \includegraphics[width=0.48\linewidth]{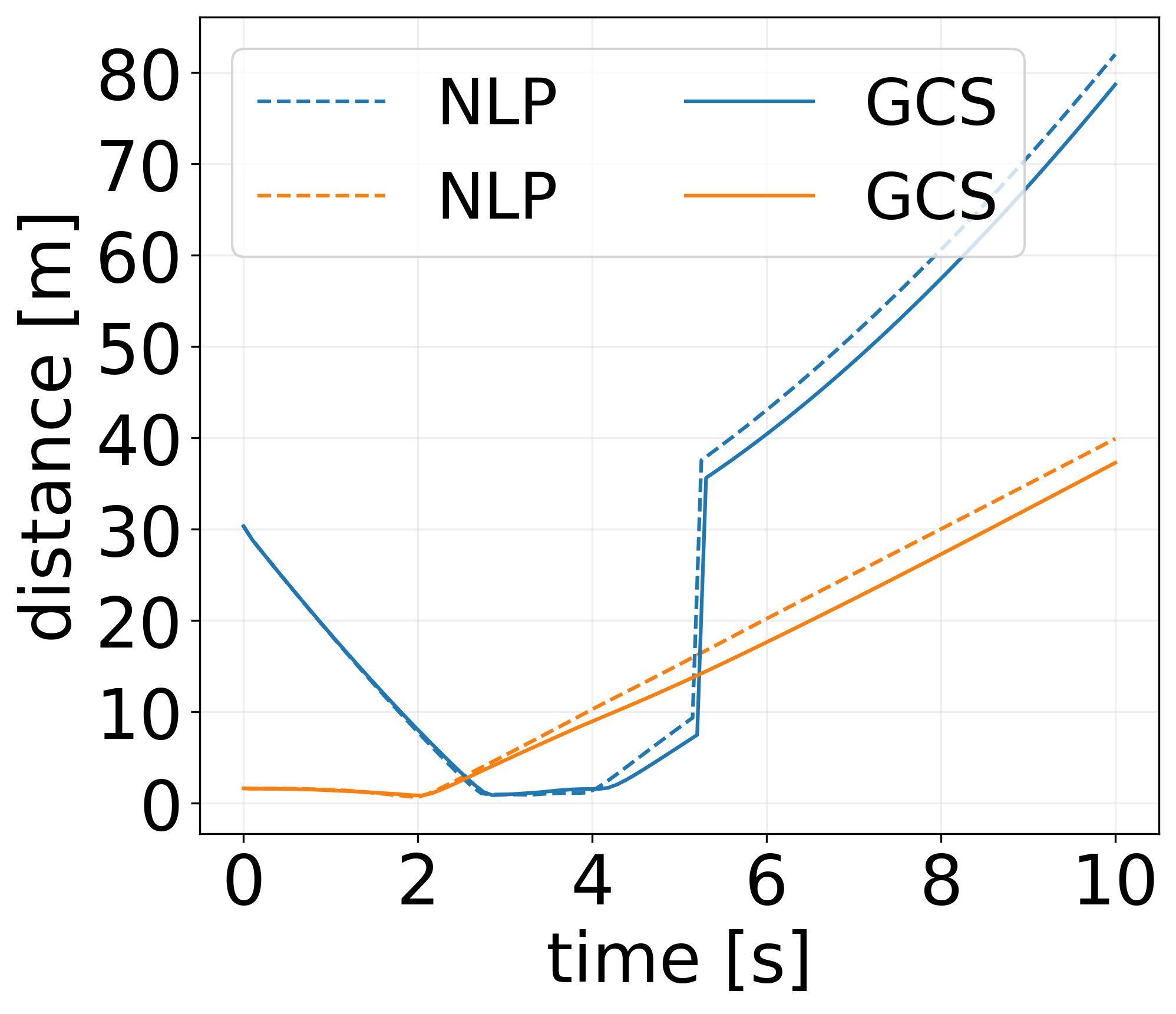}
        \label{fig:overtake:distance}
    }
     \hfill
     \subfloat[Longitudinal velocity.]{%
        \includegraphics[width=0.48\linewidth]{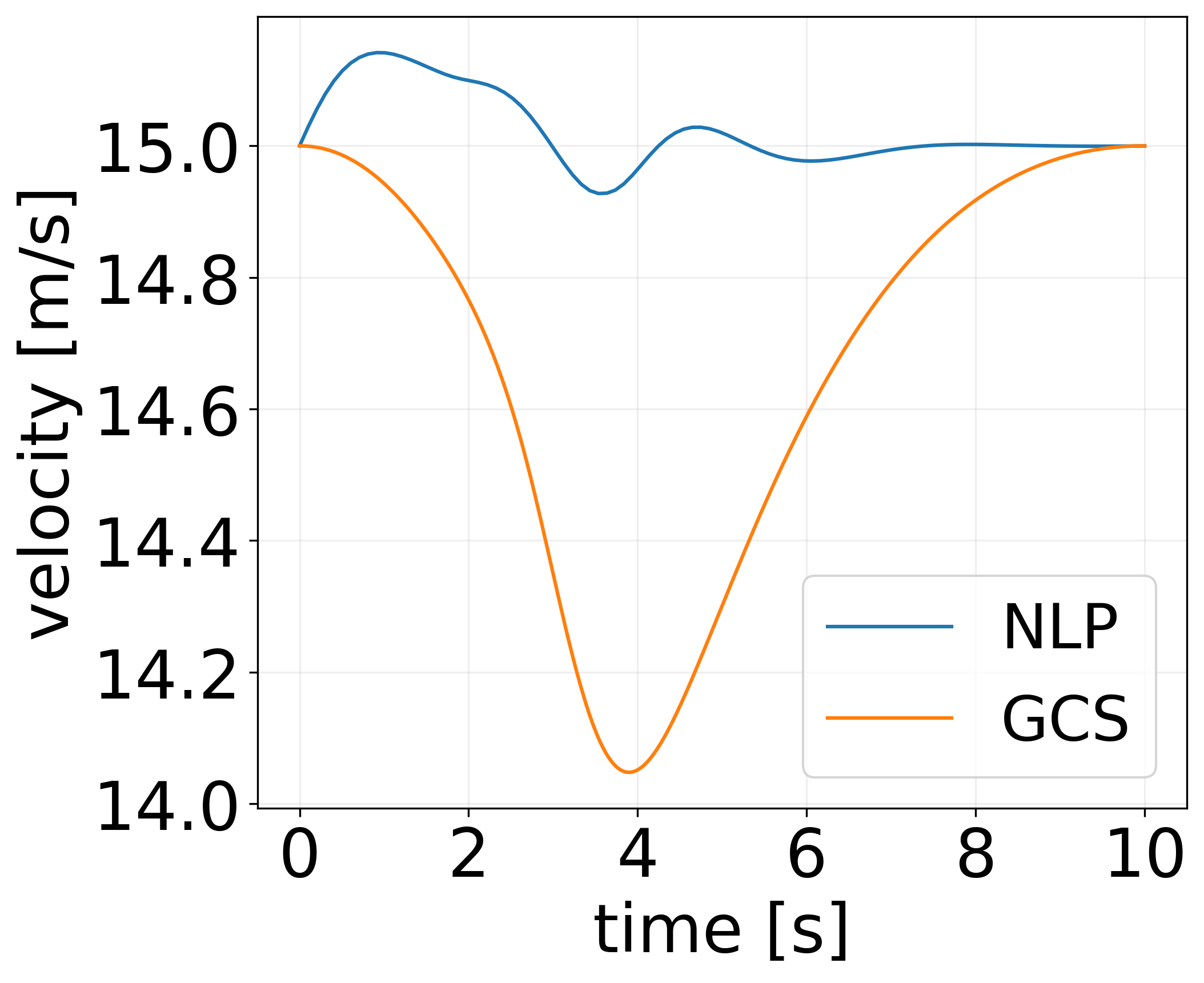}
        \label{fig:overtake:velocity}
    }
     \hfill
     \subfloat[Longitudinal acceleration.]{%
        \includegraphics[width=0.48\linewidth]{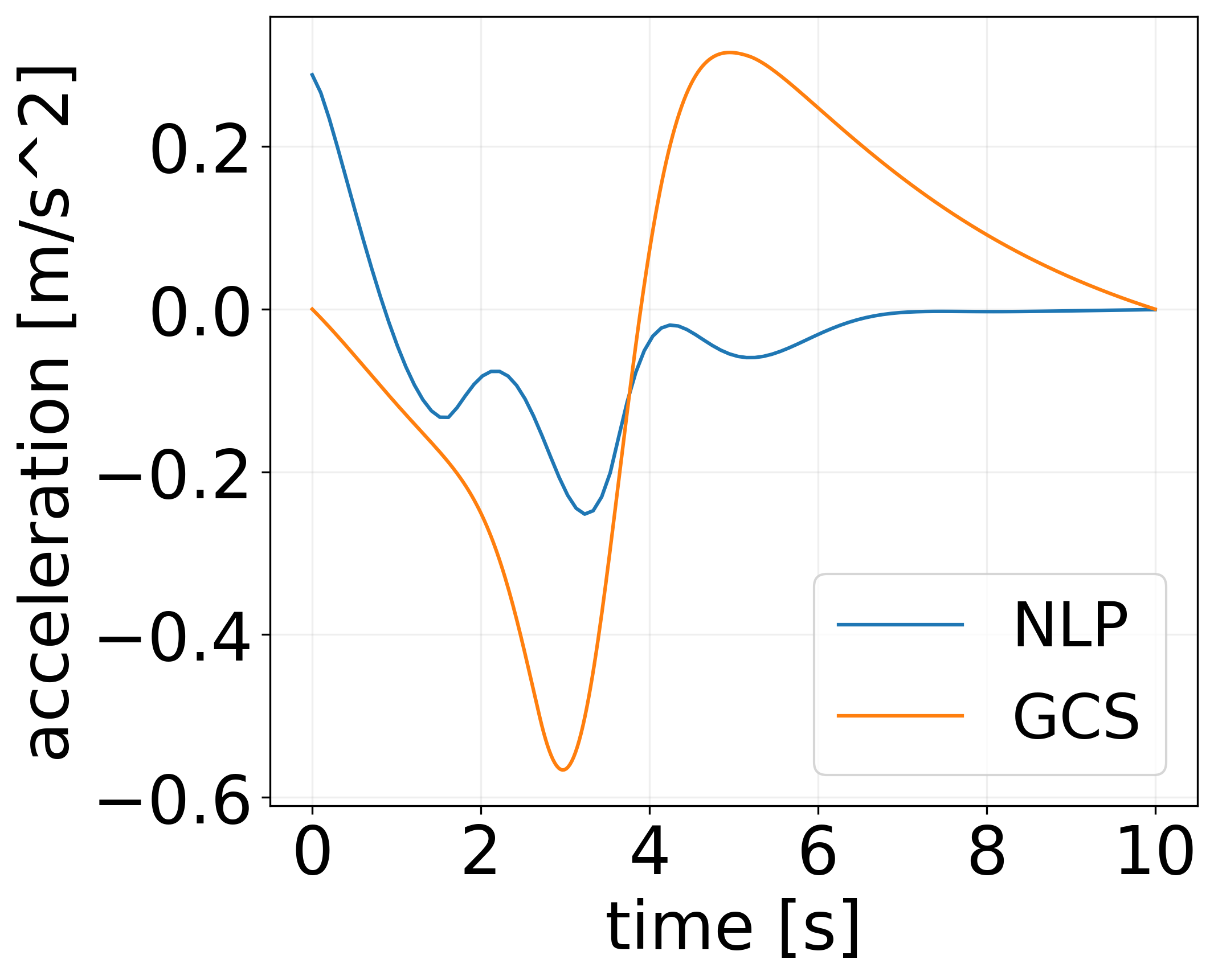}
        \label{fig:overtake:acceleration}
    }
     \hfill
     \subfloat[Steering angle.]{%
        \includegraphics[width=0.48\linewidth]{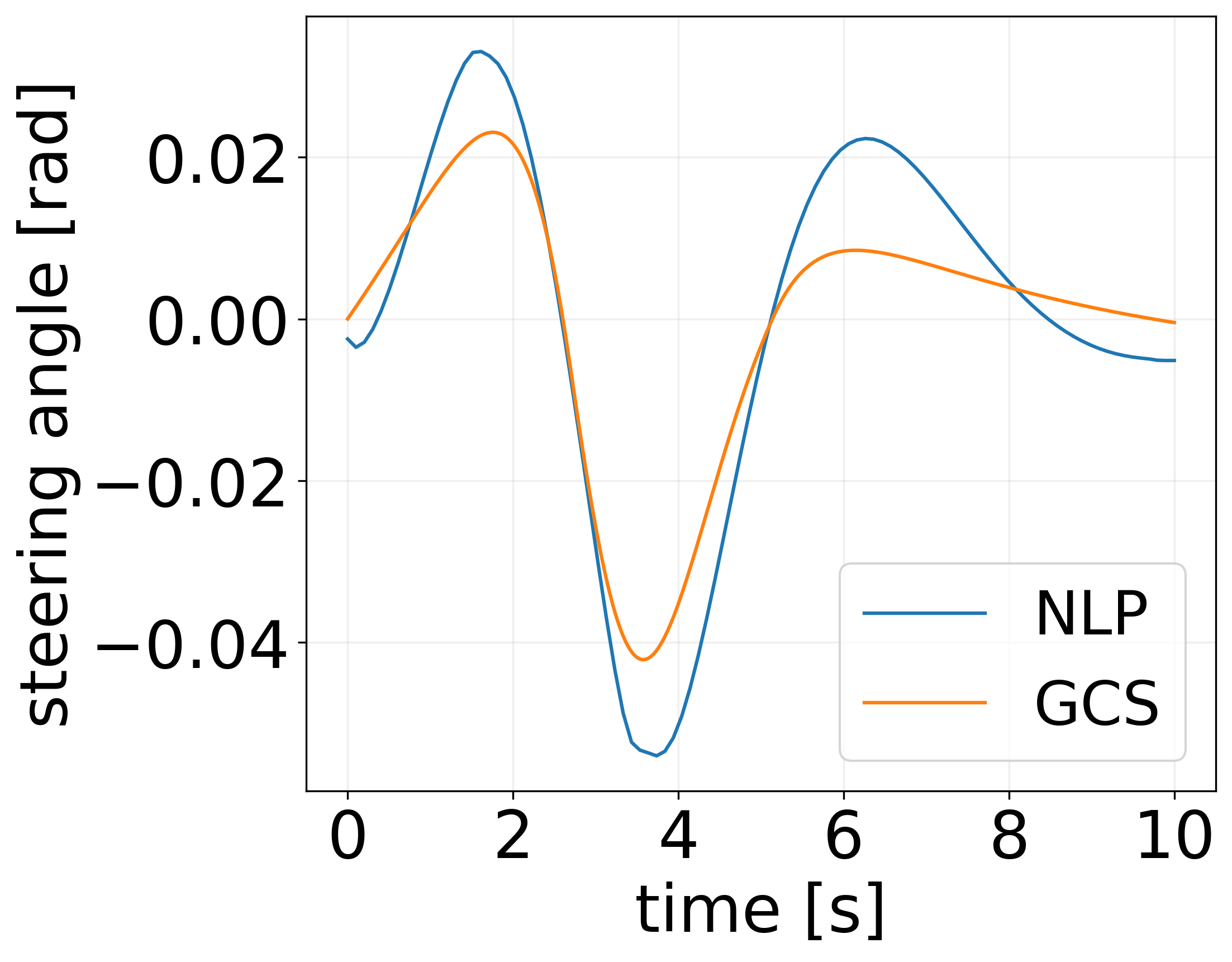}
        \label{fig:overtake:steering}
    }
    \caption{(a) Minimum distance from obstacles, (b) longitudinal velocity, (c) longitudinal acceleration and (d) steering angle associated with the trajectories obtained using \gls{GCS} and \gls{NLP} for the overtaking scenario.}
    \label{fig:overtake:curves}
\end{figure}

Figure~\ref{fig:overtake} shows the overtaking trajectories obtained using both formulations. The corresponding minimum distance, longitudinal velocity, longitudinal acceleration, and steering angle profiles are presented in Figure~\ref{fig:overtake:curves}. Both methods generate feasible overtaking maneuvers while exhibiting differences in speed and steering angles that reflect their respective optimization structures.

\subsection{Discussion}

The case studies indicate that the trajectories obtained using the \gls{GCS} and \gls{NLP} formulations are both qualitatively and quantitatively similar. Despite relying on convex free-space decomposition and polynomial trajectory parameterizations, the \gls{GCS} formulation produces solutions that roughly approximate those obtained from the nonlinear optimal control problem, which directly enforces the full vehicle dynamics and exact obstacle geometry. This suggests that, under the assumptions considered, the proposed formulation captures the dominant geometric and dynamic features relevant for motion planning.

From a computational standpoint, however, the differences are substantial. The execution times reported in Table~\ref{tab:execution time} show that the \gls{GCS} approach consistently outperforms the \gls{NLP} formulation, typically by approximately one order of magnitude. This improvement stems from the structural decomposition inherent to \gls{GCS}: combinatorial decisions are handled explicitly through graph variables, while continuous trajectory optimization within each region remains convex. As a result, the method exhibits reduced sensitivity to initialization and improved solver robustness compared to the \gls{NLP}.

Nevertheless, the dynamic profiles reveal non-negligible differences. In particular, the longitudinal acceleration in the \gls{GCS} solutions tends to be more aggressive, leading to a more oscillatory longitudinal velocity profile. This behavior is largely attributable to the difficulty of imposing tight convex constraints on acceleration within the \gls{GCS} framework. Although smoothness penalties on higher-order derivatives improve regularity, they do not fully replicate the direct nonlinear acceleration constraints available in the \gls{NLP} formulation.

One possible avenue to improve dynamic regulation is a two-stage optimization strategy, in which the spatial curve is computed first and the time-scaling is optimized subsequently. Such a decomposition would decouple path geometry from temporal scaling, potentially enabling stricter convex enforcement of acceleration bounds. The main theoretical challenge lies in guaranteeing that the spatial trajectory obtained in the first stage admits a feasible time-scaling under the imposed dynamic constraints, which remains an open research problem.

The extension of the \gls{GCS} framework to dynamic obstacle scenarios—via transition regions with associated temporal constraints—proved effective in the evaluated cases. However, the timing bounds used in these studies were derived through scenario-specific heuristic calculations. For deployment in a full autonomous driving architecture, these constraints must be generated algorithmically. In particular, a behavior planning module capable of reasoning about scene structure and predicted obstacle trajectories would be required to derive transition regions and safety timing constraints in a systematic manner. Designing this interface between behavior planning and the \gls{GCS} optimization layer constitutes an important direction for future work.

Overall, the results highlight a clear trade-off between modeling fidelity and computational structure. The \gls{GCS} formulation provides trajectories that closely approximate those of the nonlinear optimal control baseline while offering significantly improved computational efficiency and reduced sensitivity to initialization. At the same time, differences in dynamic regulation and the heuristic treatment of timing constraints indicate that further refinement is required to achieve tighter control over acceleration profiles and systematic integration with higher-level planning modules. These observations clarify both the practical advantages of the proposed approach and the key challenges that remain for its deployment in real-world autonomous driving systems.

\section{Conclusion}

This paper presented a motion planning framework for autonomous vehicles based on optimization over \gls{GCS} combined with polynomial trajectory parameterization and time scaling. The free space was decomposed into convex regions organized as a directed graph, allowing geometric nonconvexity to be treated explicitly through discrete connectivity decisions while preserving convexity of continuous trajectory constraints within each region.

Vehicle dynamics were incorporated through a simplified dynamic bicycle model under small-slip and linear tire assumptions. By exploiting the approximate flatness of the model with respect to the position coordinates, trajectory generation was performed directly in the geometric space while maintaining dynamic consistency. Bézier curves were used to parameterize spatial trajectories and time-scaling functions, enabling convex constraints on velocity, smoothness, and region containment. Continuity constraints up to third order ensured differentiability of position, velocity, and acceleration across region boundaries.

Dynamic obstacle avoidance was addressed through temporal separation constraints and velocity modulation, particularly in structured highway scenarios where geometry and timing can be partially decoupled. The resulting formulation is a finite-dimensional mixed-integer convex program whose continuous relaxation remains convex, providing computational tractability and meaningful lower bounds.

Case studies in static obstacle avoidance and lane-changing scenarios demonstrated that the proposed approach generates collision-free and dynamically feasible trajectories while exhibiting improved robustness with respect to initialization when compared to a nonlinear discrete-time optimal control baseline. 

Future work includes extending the formulation to richer vehicle models beyond the small-slip regime, incorporating tighter representations of dynamic obstacles, and investigating the separation of geometric path and time-scaling optimization as a means to obtain tighter bounds on dynamic constraints. Additionally, the integration of the proposed framework with higher-level behavioral planning strategies remains an important direction for further research, specially when it comes to improving the proposed heuristic to handle dynamic obstacles and make the computation of temporal constraints systematic.

\bibliographystyle{IEEEtran}
\bibliography{references}

\end{document}

%% file: acronyms.tex
\newacronym{GCS}{GCS}{Graphs of Convex Sets}
\newacronym{RRT}{RRT}{Rapidly Exploring Random Trees}
\newacronym{MPC}{MPC}{Model Predictive Control}
\newacronym{NLP}{NLP}{Nonlinear Program}

%% file: references.bib
@ARTICLE{zhu2025Diff,
  author={Zhu, Zicheng and Liu, Haichao and Wang, Wenxin and Duan, Jingliang and Zhao, Han and Ma, Jun},
  journal={IEEE Transactions on Intelligent Transportation Systems}, 
  title={Diffeomorphism-Transformed Iterative Linear Quadratic Regulator for Constrained Motion Planning in Autonomous Driving}, 
  year={2025},
  volume={26},
  number={10},
  pages={15175-15189},
  doi={10.1109/TITS.2025.3601596}}

@ARTICLE{botros2023Spatio,
  author={Botros, Alexander and Smith, Stephen L.},
  journal={IEEE Transactions on Intelligent Transportation Systems}, 
  title={Spatio-Temporal Lattice Planning Using Optimal Motion Primitives}, 
  year={2023},
  volume={24},
  number={11},
  pages={11950-11962},
  doi={10.1109/TITS.2023.3297068}}

@INPROCEEDINGS{zayou2025Graph,
  author={Zayou, Soufyan and Arslan, Omur},
  booktitle={2025 European Control Conference (ECC)}, 
  title={Graph-Theoretic Bézier Curve Optimization over Safe Corridors for Safe and Smooth Motion Planning}, 
  year={2025},
  volume={},
  number={},
  pages={1364-1371},
  doi={10.23919/ECC65951.2025.11187106}}

@article{Rathinam1995DifferentialFO,
  title={Differential Flatness of Mechanical Control Systems: A Catalog of Prototype Systems},
  author={Muruhan Rathinam and Richard M. Murray and Willem M. Sluis},
  journal={Dynamic Systems and Control: Volume 1 — Vibration Control; Dynamic Systems; Robotics; Sliding Mode Control; Robust and Nonlinear Control; Automated Modeling; Control of Manufacturing Processes; Precision Control},
  year={1995},
  url={https://api.semanticscholar.org/CorpusID:14167508}
}

@article{yu2021model,
  title={Model predictive control for autonomous ground vehicles: a review},
  author={Yu, Shuyou and Hirche, Matthias and Huang, Yanjun and Chen, Hong and Allg{\"o}wer, Frank},
  journal={Autonomous Intelligent Systems},
  volume={1},
  year={2021},
  publisher={Springer}
}

@article{Hu2025ARO,
  title={A Review of Learning-Based Motion Planning: Toward a Data-Driven Optimal Control Approach},
  author={Jia Hu and Yang Chang and Haoran Wang},
  journal={ArXiv},
  year={2025},
  volume={abs/2512.11944},
  url={https://api.semanticscholar.org/CorpusID:283895814}
}

@inproceedings{fan2024risk,
  title={Risk-aware self-consistent imitation learning for trajectory planning in autonomous driving},
  author={Fan, Yixuan and Li, Yali and Wang, Shengjin},
  booktitle={European Conference on Computer Vision},
  pages={270--287},
  year={2024},
  organization={Springer}
}

@inproceedings{wang2020learning,
  title={Learning hierarchical behavior and motion planning for autonomous driving},
  author={Wang, Jingke and Wang, Yue and Zhang, Dongkun and Yang, Yezhou and Xiong, Rong},
  booktitle={2020 IEEE/RSJ International Conference on Intelligent Robots and Systems (IROS)},
  pages={2235--2242},
  year={2020},
  organization={IEEE}
}

@inproceedings{yang2024diffusion,
  title={Diffusion-es: Gradient-free planning with diffusion for autonomous and instruction-guided driving},
  author={Yang, Brian and Su, Huangyuan and Gkanatsios, Nikolaos and Ke, Tsung-Wei and Jain, Ayush and Schneider, Jeff and Fragkiadaki, Katerina},
  booktitle={Proceedings of the IEEE/CVF conference on computer vision and pattern recognition},
  pages={15342--15353},
  year={2024}
}

@article{teng2023motion,
  title={Motion planning for autonomous driving: The state of the art and future perspectives},
  author={Teng, Siyu and Hu, Xuemin and Deng, Peng and Li, Bai and Li, Yuchen and Ai, Yunfeng and Yang, Dongsheng and Li, Lingxi and Xuanyuan, Zhe and Zhu, Fenghua and others},
  journal={IEEE Transactions on Intelligent Vehicles},
  volume={8},
  number={6},
  pages={3692--3711},
  year={2023},
  publisher={IEEE}
}

@article{yu2024rdt,
  title={RDT-RRT: Real-time double-tree rapidly-exploring random tree path planning for autonomous vehicles},
  author={Yu, Jiaxing and Chen, Ci and Arab, Aliasghar and Yi, Jingang and Pei, Xiaofei and Guo, Xuexun},
  journal={Expert Systems with Applications},
  volume={240},
  pages={122510},
  year={2024},
  publisher={Elsevier}
}

@article{orthey2023sampling,
  title={Sampling-based motion planning: A comparative review},
  author={Orthey, Andreas and Chamzas, Constantinos and Kavraki, Lydia E},
  journal={Annual Review of Control, Robotics, and Autonomous Systems},
  volume={7},
  year={2023},
  publisher={Annual Reviews}
}

@article{micheli2023nmpc,
  title={NMPC trajectory planner for urban autonomous driving},
  author={Micheli, Francesco and Bersani, Mattia and Arrigoni, Stefano and Braghin, Francesco and Cheli, Federico},
  journal={Vehicle system dynamics},
  volume={61},
  number={5},
  pages={1387--1409},
  year={2023},
  publisher={Taylor \& Francis}
}

@article{wang2021path,
  title={Path planning on large curvature roads using driver-vehicle-road system based on the kinematic vehicle model},
  author={Wang, Jinxiang and Yan, Yongjun and Zhang, Kuoran and Chen, Yimin and Cao, Mingcong and Yin, Guodong},
  journal={IEEE Transactions on Vehicular Technology},
  volume={71},
  number={1},
  pages={311--325},
  year={2021},
  publisher={IEEE}
}

@inproceedings{chen2014quartic,
  title={Quartic B{\'e}zier curve based trajectory generation for autonomous vehicles with curvature and velocity constraints},
  author={Chen, Cheng and He, Yuqing and Bu, Chunguang and Han, Jianda and Zhang, Xuebo},
  booktitle={2014 IEEE International Conference on Robotics and Automation (ICRA)},
  pages={6108--6113},
  year={2014},
  organization={IEEE}
}

@inproceedings{alia2015local,
  title={Local trajectory planning and tracking of autonomous vehicles, using clothoid tentacles method},
  author={Alia, Chebly and Gilles, Tagne and Reine, Talj and Ali, Charara},
  booktitle={2015 IEEE intelligent vehicles symposium (IV)},
  pages={674--679},
  year={2015},
  organization={IEEE}
}

@article{paden2016survey,
  title={A survey of motion planning and control techniques for self-driving urban vehicles},
  author={Paden, Brian and {\v{C}}{\'a}p, Michal and Yong, Sze Zheng and Yershov, Dmitry and Frazzoli, Emilio},
  journal={IEEE Transactions on intelligent vehicles},
  volume={1},
  number={1},
  pages={33--55},
  year={2016},
  publisher={IEEE}
}

@article{badue2021self,
  title={Self-driving cars: A survey},
  author={Badue, Claudine and Guidolini, R{\^a}nik and Carneiro, Raphael Vivacqua and Azevedo, Pedro and Cardoso, Vinicius B and Forechi, Avelino and Jesus, Luan and Berriel, Rodrigo and Paixao, Thiago M and Mutz, Filipe and others},
  journal={Expert systems with applications},
  volume={165},
  pages={113816},
  year={2021},
  publisher={Elsevier}
}

@article{lavalle1998rapidly,
  title={Rapidly-exploring random trees: A new tool for path planning},
  author={LaValle, Steven},
  journal={Research Report 9811},
  year={1998},
  publisher={Department of Computer Science, Iowa State University}
}

@book{betts2010practical,
  title={Practical methods for optimal control and estimation using nonlinear programming},
  author={Betts, John T},
  year={2010},
  publisher={SIAM}
}

@article{marcucci2023motion,
  title={Motion planning around obstacles with convex optimization},
  author={Marcucci, Tobia and Petersen, Mark and Von Wrangel, David and Tedrake, Russ},
  journal={Science robotics},
  volume={8},
  number={84},
  pages={eadf7843},
  year={2023},
  publisher={American Association for the Advancement of Science}
}

@article{marcucci2025unified,
  title={A Unified and Scalable Method for Optimization over Graphs of Convex Sets},
  author={Marcucci, Tobia},
  journal={arXiv preprint arXiv:2510.20184},
  year={2025}
}

@book{rajamani2011vehicle,
  title={Vehicle dynamics and control},
  author={Rajamani, Rajesh},
  year={2011},
  publisher={Springer Science \& Business Media}
}

@inproceedings{deits2015computing,
  title={Computing large convex regions of obstacle-free space through semidefinite programming},
  author={Deits, Robin and Tedrake, Russ},
  booktitle={Algorithmic Foundations of Robotics XI: Selected Contributions of the Eleventh International Workshop on the Algorithmic Foundations of Robotics},
  pages={109--124},
  year={2015},
  organization={Springer}
}

@inproceedings{CommonRoad,
	author = {Sebastian Maierhofer and Moritz Klischat and Matthias Althoff},
	title = {CommonRoad Scenario Designer: An Open-Source Toolbox for Map Conversion and Scenario Creation for Autonomous Vehicles},
	booktitle = {Proc. of the IEEE Int. Conf. on Intelligent Transportation Systems },
	year = {2021},
    pages = {3176-3182},
    doi = {10.1109/itsc48978.2021.9564885}
}

@inproceedings{wagner2023rti,
  title={RTI-NMPC for Control of Autonomous Vehicles Using Implicit Discretization Methods},
  author={Wagner, Matheus and Normey-Rico, Julio E},
  booktitle={Simp{\'o}sio Brasileiro de Automa{\c{c}}{\~a}o Inteligente-SBAI},
  volume={1},
  number={2},
  year={2023}
}
